\RequirePackage{fix-cm}

\documentclass[twocolumn]{svjour3}          % twocolumn
\smartqed  % flush right qed marks, e.g. at end of proof
\usepackage{graphicx}
\usepackage{amsmath}
\usepackage{subcaption}
\usepackage{mathabx}
\usepackage[bookmarks=true]{hyperref}
\usepackage[numbers]{natbib}
\usepackage{tikz}
\usepackage{xcolor}

\journalname{Computing and Visualization in Science}

\begin{document}

\title{Parareal with a Learned Coarse Model \\
for Robotic Manipulation \thanks{This project was
supported by an EPSRC studentship (1879668) and EPSRC grants EP/R031193/1 and EP/P019560/1.}
}

\author{Wisdom Agboh \and Oliver Grainger \and Daniel Ruprecht \and Mehmet Dogar}

\institute{
            W. Agboh \at
            School of Computing, University of Leeds, United Kingdom. \email{w.c.agboh@leeds.ac.uk}
            \and
            O. Grainger \at
            School of Mechanical Engineering, University of Leeds, United Kingdom. \email{mn17omg@leeds.ac.uk}
            \and
            D. Ruprecht \at
            Lehrstuhl Computational Mathematics, Institut für Mathematik, Technische Universität Hamburg, Germany. \email{ruprecht@tuhh.de}
            \and
            M. Dogar \at
            School of Computing, University of Leeds, United Kingdom. \email{m.r.dogar@leeds.ac.uk}
}

\date{Received: date / Accepted: date}

\maketitle
\begin{abstract}
A key component of many robotics model-based planning and control algorithms is physics predictions, that is, forecasting a sequence of states given an initial state and a sequence of controls.
This process is slow and a major computational bottleneck for robotics planning algorithms.
Parallel-in-time integration methods can help to leverage parallel computing to accelerate physics predictions and thus planning.

The Parareal algorithm iterates between a coarse serial integrator and a fine parallel integrator.
A key challenge is to devise a coarse model that is computationally cheap but accurate enough for Parareal to converge quickly.
Here, we investigate the use of a deep neural network physics model as a coarse model for Parareal in the context of robotic manipulation.

In simulated experiments using the physics engine Mujoco as fine propagator we show that the learned coarse model leads to faster Parareal convergence  than a coarse physics-based model.
We further show that the learned coarse model allows to apply Parareal to scenarios with multiple objects, where the physics-based coarse model is not applicable.

Finally, we conduct experiments on a real robot and show that Parareal predictions are close to real-world physics predictions for robotic pushing of multiple objects. { Code\footnote{\url{https://doi.org/10.5281/zenodo.3779085}} and videos\footnote{\url{https://youtu.be/wCh2o1rf-gA}} are publicly available. }

\end{abstract}

\section{Introduction}
\label{sec:introduction}

\begin{figure*}[htb!]
  \centering
  \begin{subfigure}[b]{0.245\textwidth}
  \centering
     \includegraphics[scale=0.100]{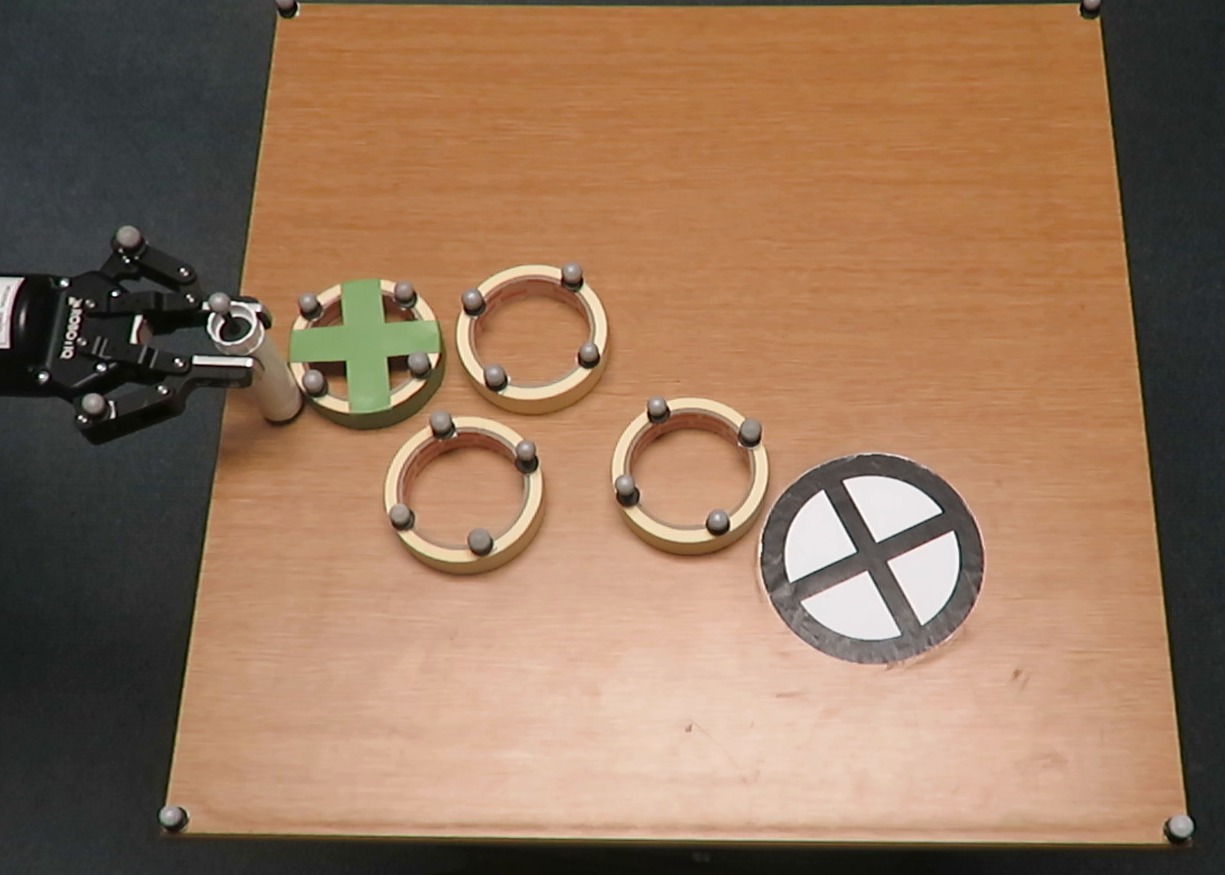}
    \end{subfigure}
  \begin{subfigure}[b]{0.245\textwidth}
  \centering
    \includegraphics[scale=0.100]{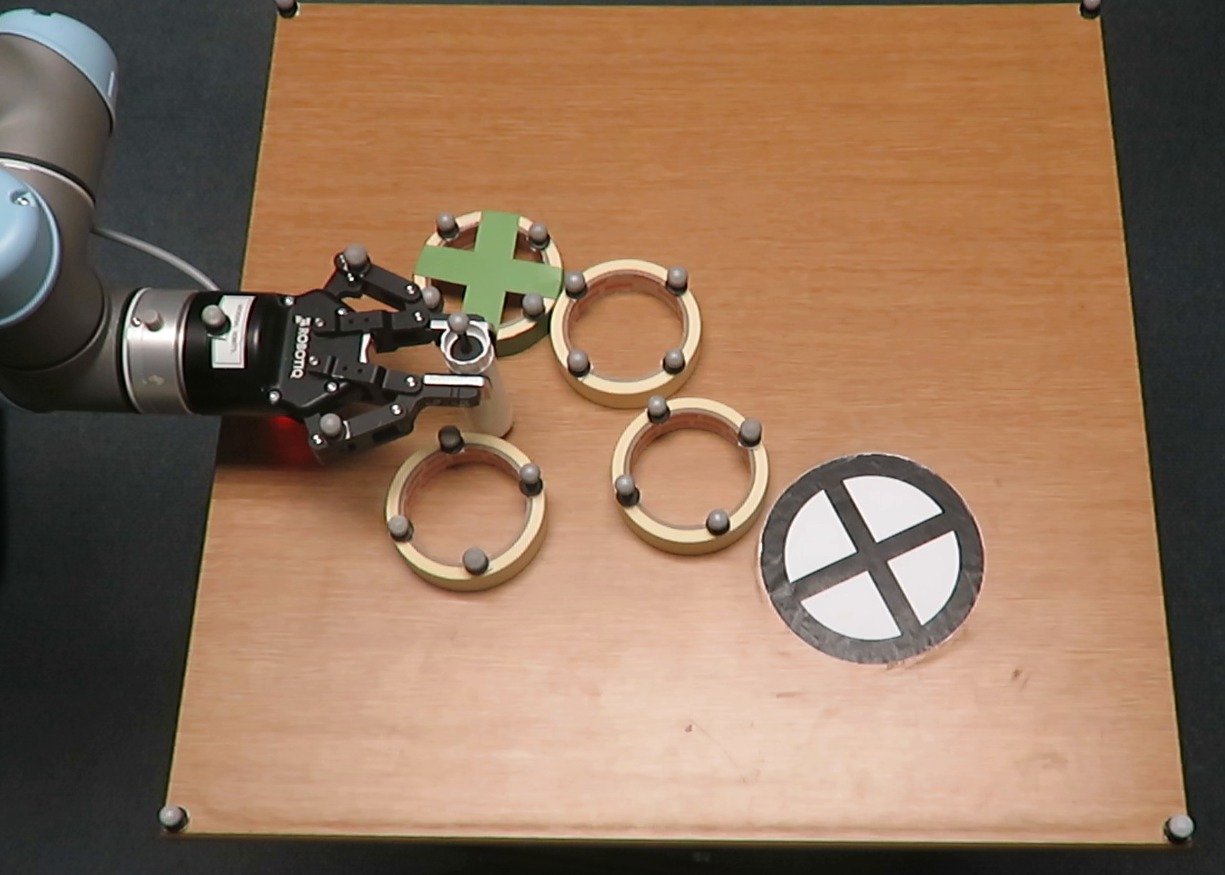}
  \end{subfigure}
     \begin{subfigure}[b]{0.245\textwidth}
    \includegraphics[scale=0.100]{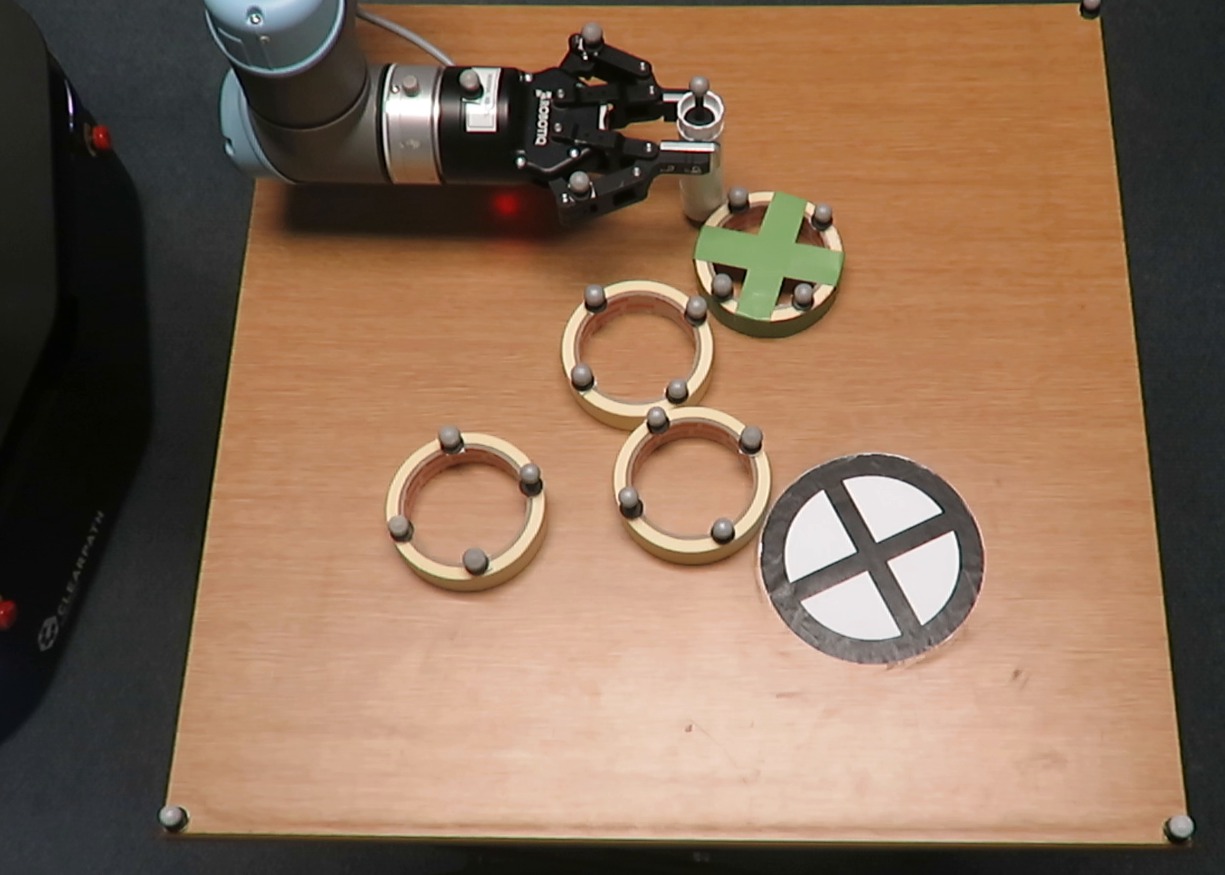}
  \end{subfigure}
  \begin{subfigure}[b]{0.245\textwidth}
    \includegraphics[scale=0.100]{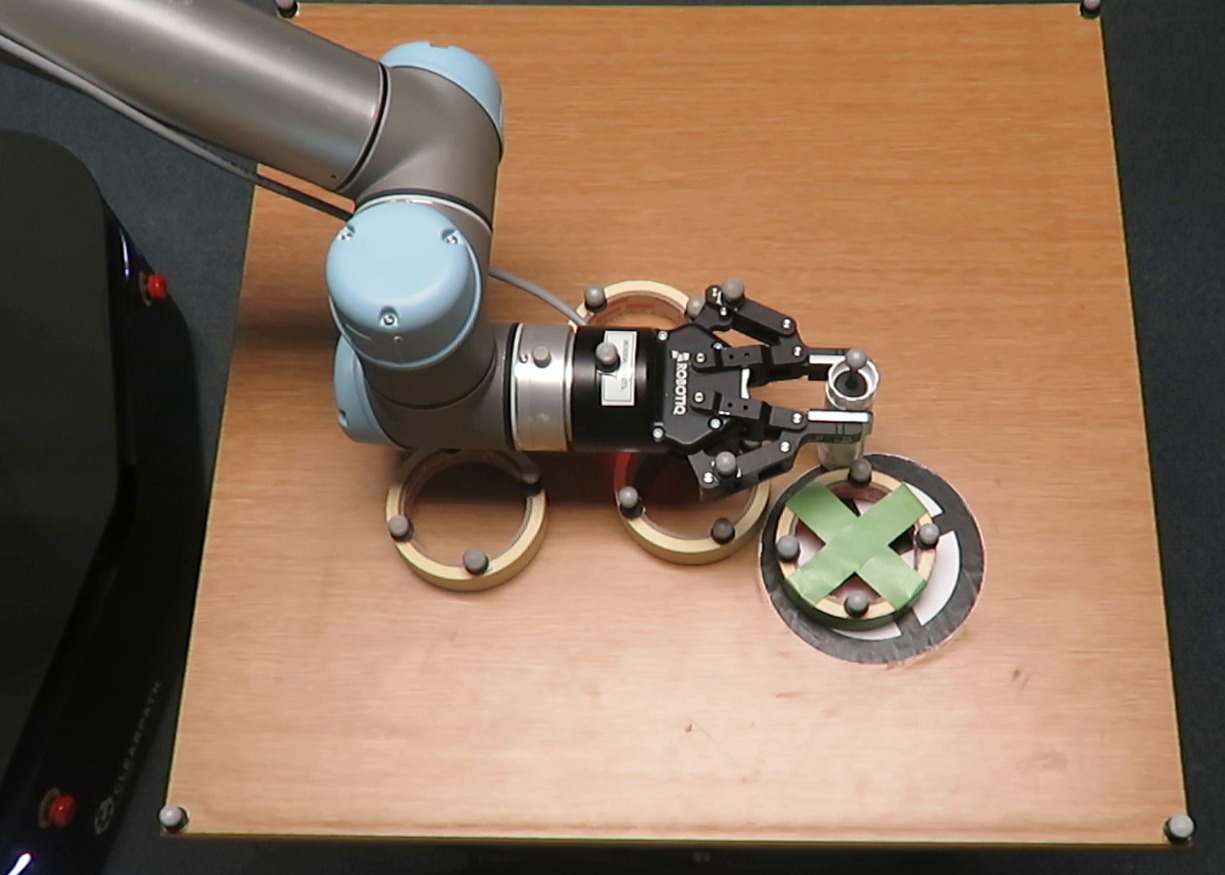}
  \end{subfigure}
  \caption{Example of a robotic manipulation planning and control task using physics predictions. The robot controls the motion of the green object solely through contact. The goal is to push the green object into the target region marked $X$. The robot must complete the task without pushing other objects off the table or into the goal region.}
  \label{fig:robot_manipulation}
\end{figure*}
We present a method for fast and accurate physics predictions
during non-prehensile manipulation planning and control.
An example scenario is shown in Figure~\ref{fig:robot_manipulation}, where a robot arm pushes the marked cylindrical object into a target zone without pushing the other three objects off the table.
We are interested in predicting the motion of the objects in a fast and accurate way.

Physics engines like Mujoco {\cite{mujoco}} and Drake~\cite{drake} solve Newton's equation to predict motion.
They are accurate but slow.
Coarse models can be built by introducing simplifying assumptions, trading accuracy for solution speed but their lack of precision will eventually compromise the robot's chance of completing a given task successfully.

Given an initial state and a sequence of controls, the problem of predicting the resulting sequence of states is a key component of a variety of model-based planning and control algorithms~\citep{pusher_slider,kalakrishnan2011stomp,convergent_planning,mppi}. Mathematically, such a prediction requires solving an initial value problem.
Typically, those are solved through numerical integration over time-steps using e.g. semi-implicit Euler's method or Runge-Kutta methods and an underlying physics model to provide the forces.
However, the speed with which these accurate physics-based predictions can be performed is still slow~\citep{compare_engine_ErezICRA2015}. Faster physics-based predictions can contribute significantly to contact-based/non-prehensile manipulation planning and control {--- especially during re-planning or model-predictive control (MPC) where a robot executes an action in the real-world, gets the resulting state and then has to generate a new physics-based plan.} Such MPC methods have been used in prior work to achieve  manipulation robustness to parameter uncertainty \cite{mpc_parameter_uncertainty}, stabilize complex humanoid behaviours \cite{tassa_synthesis}, and visually manipulate fabric \cite{visualmpc}.

In a previous paper~\cite{agboh_isrr19}, we demonstrated that predictions for a robot pushing a single object can be made faster by combining a fine physics-based model with a simple, coarse physics-based model using the parallel-in-time method Parareal.
Using 4 cores, Parareal was about a factor two faster than the fine physics engine alone while providing comparable accuracy and the same success rate for a push planning problem with obstacle avoidance.
Here, we extend these results by investigating a deep neural network as coarse model and show that it leads to faster Parareal convergence.
We also demonstrate that Parareal can be used to speed up physics prediction in scenarios where the robot pushes multiple objects.
\section{Related Work}
\label{sec:related_work}
Parareal has been used in many different areas.
Trindade et al., for example, use it to simulate incompressible laminar flows~\citep{parareal_fluidsim}.
Maday et al. have tested it for to simulate dynamics in quantum chemistry~\citep{parareal_quantum_control}.
The method was introduced by Lions et al. in 2001~\cite{parareal_Lions}.
Combinations of parallel-in-time integration and neural networks have not yet been studied widely.
Very recently, Yalla and Enquist showed the promise of using a machine learned model as coarse propagator~\cite{YallaEnquist2018} for test problems.
Going the other way, Schroder~\cite{Schroder2017} and Günther et al.~\cite{GuentherEtAl2019} recently showed that parallel-in-time integration can be used to speed up the process of training neural networks.

{Results on how Parareal performs for differential algebraic equations (DAEs) are scarce.
Guibert et al.~\cite{GuibertTromeur2007} demonstrate that Parareal can solve DAEs, but can experiences issues with stability for very stiff problems.
Cadeau et al.~\cite{Cadeau2011} propose a combination of Parareal with waveform relaxation to introduce additional parallelism.
For a DAE system of size 100,000, they demonstrate that adding Parareal does  provide speedup beyond the saturation point of waveform relaxation alone.}

Physics predictions play a major role in robotic manipulation planning and control -- to generate uncertainty averse robotic pushing plans \citep{mppi_push}, to manipulate objects in clutter through online re-planning \cite{agboh_humanoids18}, to rearrange objects in clutter through dynamic actions \citep{haustein_asfour}, and also to use human guidance to generate pushing motions~\cite{papallasicra2020}. However, planning is slow since physics predictions  are computationally expensive. Parareal's potential to speed up simulations for robotic manipulation in single-object scenarios using a physics-based coarse model was recently demonstrated by Agboh et al.~\cite{agboh_isrr19}.

Furthermore, physics predictions are essential in learning physics-based manipulation policies. For example, learning gentle object manipulation through curiosity~\cite{gentle_object_manip}, learning long-horizon robotic agent behaviours through latent imagination~\cite{dreamer}, learning visuo-motor policies by formulating exploration as a latent trajectory optimization problem \cite{latent_traj_opt}, learning policies for manipulation in clutter\cite{wissam_iros19}, smoothing fabric with a da Vinci surgical robot through deep imitation learning \cite{seita_fabrics_2019}, and learning human-like manipulation policies through virtual reality demonstrations \cite{hasan2020humanlike}. The training time for these policies can potentially be reduced with a parallel-in-time approach to physics predictions.

Combining different physics models for robotic manipulation has been the topic
of recent research, although not with a focus on improving prediction speed.
Kloss et al.~\cite{combining_learned_analytical_Kloss_C0RR_17}
address the question of accuracy and generalization in combined
neural-analytical models. Ajay et al.~\cite{augmenting_physics_Ajay_IROS18} focus on modeling the inherent stochastic nature of the real world physics, by combining an analytical, deterministic rigid-body simulator with a stochastic neural network.

We can make physics engines faster by using larger simulation time steps. However, this decreases the accuracy and can result in unstable behavior where objects have unrealistically large accelerations.
To generate stable behaviour at large time-step sizes, Pan et al.~\cite{position_based_collocation_Pan_WAFR18} propose an integrator for articulated body dynamics by using only position variables to formulate the
dynamic equation.
Moreover, Fan et al.~\cite{Variational_Integrators_Fan_WAFR18} propose linear-time variational integrators of arbitrarily high order for robotic simulation and use them in trajectory optimization to complete robotics tasks.
Recent work has focused on making the underlying planning and control
algorithms faster.
For example, Giftthaler et al.~\cite{GNMS_Giftthaler_CoRR17} introduced a multiple-shooting variant of the trajectory optimizer - iterative linear quadratic regulator \cite{ilqr} which has shown impressive results for real-time
nonlinear optimal control of complex robotic systems.
\citep{whole_body_mpc_buchli_RAL18, performance_DDP_Plancher_WAFR18}.

\section{Robotic Manipulation with Parareal}
\subsection{Robotic manipulation}\label{sec:robotic_manipulation}
Consider the scene shown in Figure~\ref{fig:robot_manipulation}.
The robot's manipulation task is to control the motion of the green goal object through pushing contact from the cylindrical pusher in the robot's gripper.
The robot needs to push the goal object into a goal region marked with an $X$.
It is allowed to make contact with other sliders but not to push them off the table or into the goal region.

The system's state at time point $n$ consists of the pose $\vec{q}$ and velocities, $\dot{\vec{q}}$ of the pusher $P$ and $N_s$ sliders, $S^{i} \dots S^{N_s}$:
\begin{align*}
    \vec{x}_{n} = [\vec{q}^{P}_{n}, \vec{q}^{S^{i}}_{n}, \dots ,\vec{q}^{S^{N_s}}_{n}, \dot{\vec{q}}^{P}_{n}, \dot{\vec{q}}^{S^{i}}_{n}, \dots ,\dot{\vec{q}}^{S^{N_s}}_{n}].
\end{align*}

The pose of slider $i$ consists of its position and orientation on the plane: ${\vec{q}^{S^{i}}=[q^{S^{i}_x},q^{S^{i}_y},q^{S^{i}_{\theta}}]^T}$. The pusher's pose is ${\vec{q}^{P}=[q^{P_x},q^{P_y}]^T}$ and control inputs are velocities ${\vec{u}_{n} = [u^{x}_{n}, u^{y}_{n}]^T}$ applied on the pusher at time $n$ for a  control duration of $\Delta t$.

A robotics planning and control algorithm takes in an initial state of the system $\vec{x}_{0}$, and outputs an optimal sequence of controls $\{\vec{u}_{0}, \vec{u}_{1}, \dots, \vec{u}_{N-1} \}$. However, to generate this optimal sequence, the planner needs to simulate many different control sequences and predict many resulting sequences of states ${\{\vec{x}_{1}, \vec{x}_{2}, \dots, \vec{x}_{N} \}}$.

The planner makes these simulations through a physics model $F$ of the real-world that predicts the next state $\vec{x}_{n+1}$ given the current state $\vec{x}_{n}$ and a control input $\vec{u}_{n}$
\begin{align} \label{eq:dynamics}
    \vec{x}_{n+1} = F(\vec{x}_n,\vec{u}_n,\Delta t).
    \vspace{-2mm}
\end{align}
 We use the general physics engine Mujoco~\citep{mujoco} to model $F$.  {It solves differential algebraic equations of motion for the complex multi-contact dynamics problem
 \begin{align} \label{eq:mujoco_equations}
     \begin{split}
     M(\vec{q}) \hspace{1mm} d\vec{v} = (\vec{b}(\vec{q}, \vec{v}) + \tau ) \hspace{1mm} dt + J_{E}(\vec{q})^{T}\vec{f}_{E}(\vec{q}, \vec{v}, \tau) \\
     \hspace{-5mm} + J_{C}(\vec{q})^{T}\vec{f}_{C}(\vec{q}, \vec{v},\tau)
     \end{split}
 \end{align}
 where \vec{q}, \vec{v}, and M are position vector, velocity vector, and inertia matrix respectively in generalized coordinates. \vec{b} contains bias forces (Coriolis, gravity, centrifugal, springs), $\vec{f}_{E}$ and $\vec{f}_{C}$ are impulses caused by equality constraints and contacts respectively and $J_{E}$ and $J_{C}$ are the corresponding Jacobians and $\tau$ are external/applied forces. The equations are then solved numerically. Mujoco obtains a discrete-time system with two options for integrators --- semi-implicit Euler or $4^{th}$ order explicit Runge-Kutta.
 }

\subsection{Parareal}
Normally, computing all states $\vec{x}_n$ happens in a serial fashion, by evaluating~\eqref{eq:dynamics} first for $n=0$, then for $n=1$, etc.
Parareal replaces this inherently serial procedure by a parallel-in-time integration process where some of the work can be done in parallel.
For Parareal, we need a coarse physics model
%%%
\begin{equation}
    \label{eq:coarse_dynamics}
    \vec{x}_{n+1} = C(\vec{x}_n,\vec{u}_n,\Delta t).
    \vspace{-2mm}
\end{equation}
%%%
It needs to be computationally cheap relative to the fine model but does not have to be very accurate.
Parareal begins by computing an initial guess  $\vec{x}^{k=0}_n$  of the state at each time point $n$ of the trajectory using the coarse model.

This guess is then corrected via the Parreal iteration
\begin{equation}
    \label{eq:parareal}
    \vec{x}^{k+1}_{n+1} = C(\vec{x}^{k+1}_n, \vec{u}_n, \Delta t) + F(\vec{x}^{k}_n, \vec{u}_n, \Delta t) - C(\vec{x}^k_n, \vec{u}_n, \Delta t),
\end{equation}
for all timesteps ${n=0,\dots,N-1}$.
The newly introduced superscript $k$ counts the number of Parareal iterations.
The key point in iteration (\ref{eq:parareal}) is that evaluating the fine physics model can be done in parallel for all $n=0, \ldots, N-1$, while only the fast coarse model has to be computed serially.

After one Parareal iteration, $\vec{x}^{1}_{1}$ is exactly the fine solution. After two iterations, $\vec{x}^{1}_{1}$ and $\vec{x}^{2}_{2}$ are exactly the fine solutions. When $k=N$, Parareal produces the exact fine solution~\cite{parareal_analysis_martin,parareal_Lions}.
However, to produce speed up, we need to stop Parareal at much earlier iterations.
This way, Parareal can run in less wall-clock time than running the fine model serially step-by-step.
Below, we demonstrate that even after a small number of iterations, the solution produced by Parareal is of sufficient quality to allow our robot to succeed with different tasks.
Note that, for the sake of simplicity, we assume here that the number of controls $N$ and the number of processors used to parallelize in time are identical, but this can easily be generalised.

\section{Coarse models}
In this section, we introduce two coarse physics models for Parareal - a learned coarse model and the analytical coarse model from Agboh et al.~\cite{agboh_isrr19}.
\subsection{Learned coarse model}
As an alternative to the coarse physics model, we train a deep neural network as a coarse model for Parareal for robotic pushing.

\subsubsection{Network architecture}
The input to our neural network model is a state $\vec{x}_{n}$ and a single action $\vec{u}_{n}$.
{The output is the change in state $\Delta\vec{x}$ which is added to the input state to obtain the next state $\vec{x}_{n+1}$}. We use a feed-forward deep neural network (DNN) with 5 fully connected layers.
The first 4 contain 512, 256, 128 and 64 neurons, respectively, with ReLU activation function.
The output layer contains 24 neurons with linear activation functions.

\subsubsection{Dataset}
We collect training data using the physics engine Mujoco~\cite{mujoco}.
Each training sample is a tuple ($\vec{x}_{n}, \vec{u}_{n}, \vec{x}_{n+1}$).
It contains a randomly\footnote{{We use rejection sampling to ensure that sampled states do not have objects in penetration, i.e. fulfill the algebraic constraints of Eq.~\ref{eq:mujoco_equations}. }} sampled initial state, action, and next state.
We collect over 2 million such samples from the physics simulator.

During robotic pushing, a physics model may need to predict the resulting state even for cases when there is no contact between pusher and slider.
We include both contact and no-contact cases in the training data.

We train a single neural network to handle one pusher with at least one and at most $N_s$ objects being pushed (also called sliders).
While collecting data for a particular number of sliders, we placed the unused sliders in distinct fixed positions outside the pushing workspace.
These exact positions must be passed to the neural network at test time if fewer than $N_{s}$ sliders are active. For example, if $N_s=4$, to make a prediction for a 3 slider scene, we place the last slider at the same fixed position used during training.
\subsubsection{Loss function}
% OLIVER - I think it would be clearer to remove the batch notation entirely, and just show the loss function as if it were for a single sample, which would then be assumed to be averaged over a batch of samples. The original equation was not consistent in that it did not sum and average over the batch for all terms. I have commented the old equation and rewritten a new one.
The standard loss function for training is the mean squared error between the network's prediction and the training data.
On its own, this leads to infeasible state predictions where there is pusher-slider or slider-slider penetration.
We resolve this by adding a no penetration loss term such that the final loss function reads:
%
%\begin{align} \label{eq:loss_function}
%\begin{split}
%f_{l} = \frac{1}{B \cdot V}\sum_{i=1}^{N_s} \sum_{j=1}^{N_s} \cdot %{||\vec{x}^{f}_{ij} - \vec{x}^{NN}_{ij}||^{2}} \\
%+ W_{F} \cdot \sum_{i=1}^{N_s} \min(||\vec{q}^{p} - \vec{p}_{i}|| - (r_{p} + %r_{i}), 0)^{2} \\
%+ W_{F} \cdot \sum_{i=1}^{N_s} \sum_{j=i+1}^{N_s} \min(||\vec{p}_{i} - %\vec{p}_{j}|| - (r_{i} + r_{j}), 0)^{2}.
%\end{split}
%\end{align}
%
\begin{align} \label{eq:loss_function}
\begin{split}
{
f_{l} = W_{F} \cdot \sum_{i=1}^{N_s} \sum_{j=i+1}^{N_s} \min(||\vec{p}^{NN}_{i} - \vec{p}^{NN}_{j}|| - (r_{i} + r_{j}), 0)^{2}} \\ {
+ W_{F} \cdot \sum_{i=1}^{N_s} \min(||{\vec{p}_{P}} - \vec{p}^{{NN}}_{i}|| - (r_{p} + r_{i}), 0)^{2} }\\
{+ {||\vec{x}^{f} - \vec{x}^{NN}||^{2}}}.
\end{split}
\end{align}
%
% Wisdom: Position of pusher has been defined previously in 3.1, I suggest we remain consistent in terms of notation.
Here, $W_{F}$ is a constant weight, $\vec{x}^{f}$ is the next state predicted by the fine model, $\vec{x}^{NN}$ is the next state predicted by the DNN model. $\vec{p}^{NN}_{i}$ and $\vec{p}^{NN}_{j}$ are the new positions of sliders $i$ and $j$ predicted by the DNN model, respectively, and $\vec{p}_{P}$ is the position of the pusher. $r_{p}$ is the radius of the pusher, and $r_{i}$, $r_{j}$ represent the radius of sliders $i$ and $j$, respectively.
The first line of Equation~\ref{eq:loss_function} penalizes slider-slider penetration,
the second line penalizes pusher-slider penetration, and the third line is the standard mean squared error.

Finally, the network makes a single step prediction. However, robotic manipulation typically needs a multi-step prediction as a result of a control sequence. To do this, we start from the initial state and apply the first action in the sequence to get a resulting next state. Then, we use this next state as a new input to the network together with the second action in the sequence and so on. This way, we repeatedly query the network with its previous predictions as the current state input.

\subsection{Analytical coarse model}
Agboh et al.~\cite{agboh_isrr19} have proposed a simple, kinematic coarse physics model  for pushing a single object.
The model moves the slider with the same linear velocity as the pusher as long as there is contact between the two.
We give details below for completeness:

\begin{align}\label{eq:coarse_slider_pose_update}
\centering
     \vec{q}^{S}_{n+1} = \vec{q}^{S}_{n} + [u^x_n, u^y_n, \omega]^T \cdot p_c \cdot \Delta t
\end{align}
\begin{align}
    p_{c} = \frac{d_{contact}}{d_{contact} + d_{free}}, \hspace{5mm} \omega = K_{\omega} \cdot \frac{||\vec{u}_{n}|| \cdot \sin{\theta}}{||\vec{r}_{c}||}
\end{align}
\begin{align}\label{eq:coarse_slider_vel_update}
     \dot{\vec{q}}^{S}_{n+1} = \{ [u^x_n, u^y_n, \omega]^T
     \hspace{2mm} if \hspace{2mm} p_c > 0, \hspace{2mm} \dot{\vec{q}}^{S}_{n} \hspace{2mm} otherwise \}
\end{align}
\begin{align}\label{eq:coarse_pusher_update}
    \vec{q}^{P}_{n+1} = \vec{q}^{P}_{n} + \vec{u}_{n} \cdot \Delta t, \hspace{5mm} \dot{\vec{q}}^{P}_{n+1} = \vec{u}_{n}.
\end{align}

Here, $p_{c}$ is the ratio of contact distance $d_{contact}$ travelled by the pusher when in contact with the slider and the total pushing distance, $\vec{r}_c$ is a vector from the contact point to the object's center at the current state $\vec{q}^{S}_{n}$,  $\theta$ is the angle between the pushing direction and the vector $\vec{r}_c$, $\omega$ is the coarse angular velocity induced by the pusher on the slider. $K_{\omega}$ is a positive constant.

\section{Planning and control}

We use the predictive model based on Parareal described above in a planning and control framework for pushing an object on a table to a target location.
We take an optimization approach to solve this problem.
Given the table geometry, goal position, the current state of the pusher and all sliders $\vec{x}_0$, and an initial candidate sequence of controls ${\{\vec{u}_0,\vec{u}_1,\dots,\vec{u}_{N-1}\}}$, the optimization procedure outputs an optimal sequence ${\{\vec{u}^*_0,\vec{u}^*_1,\dots,\vec{u}^*_{N-1}\}}$ according to some defined cost.

The predictive model is used within this optimizer to \textit{roll-out} a sequence of controls to predict the states ${\{\vec{x}_1, \dots, \vec{x}_{N}\}}$.
These are then used to compute the cost associated with those controls.
The details of the exact trajectory optimizer can be found in Agboh et al.~\cite{agboh_wafr18}.
The cost function we use penalizes moving obstacle sliders and dropping objects from the table but encourages getting the goal object into the goal location.

We use the trajectory optimizer in a model-predictive control (MPC) framework. Once we get an output control sequence from the optimizer, we \textit{do not} execute the whole sequence on the real-robot serially one after the other.
Instead, we execute only the first action, update $\vec{x}_0$ with the observed state of the system, and repeat the optimization to generate a new control sequence.
We repeat this process until the task is complete.

Such an optimization-based MPC approach to pushing manipulation is frequently used to handle uncertainty and improve success in the real-world~\citep{agboh_humanoids18,mppi_push,pusher_slider,combining_learned_analytical_Kloss_C0RR_17}.
Here, our focus is to evaluate the performance of Parareal with learned coarse model for planning and control.

\section{Experiments and Results}
\label{sec:experiments_results}
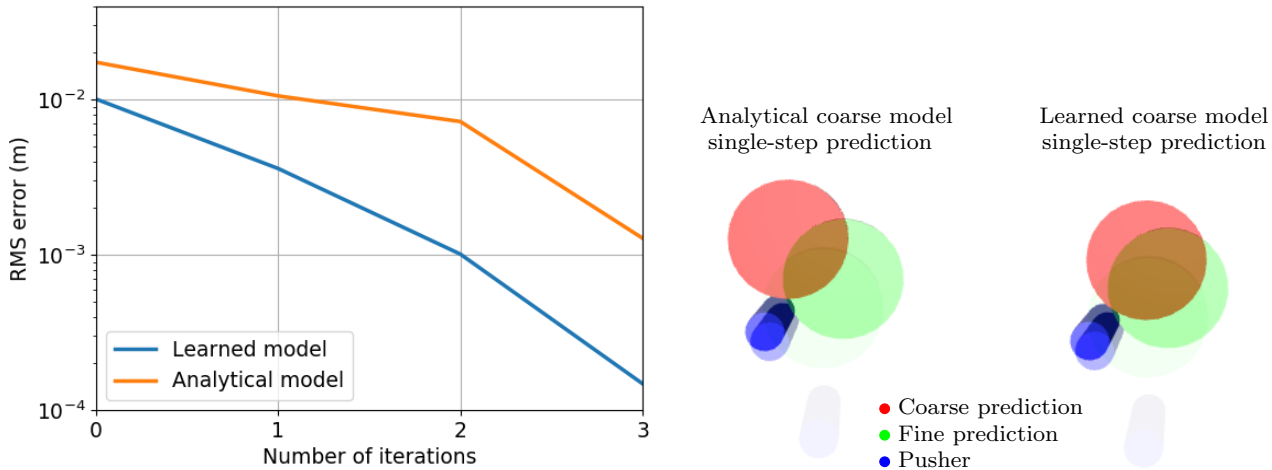
\begin{figure*}[htb!]
	\centering
	\begin{subfigure}[b]{0.49\textwidth}\label{fig:single_object_error}
		\includegraphics[scale=0.57]{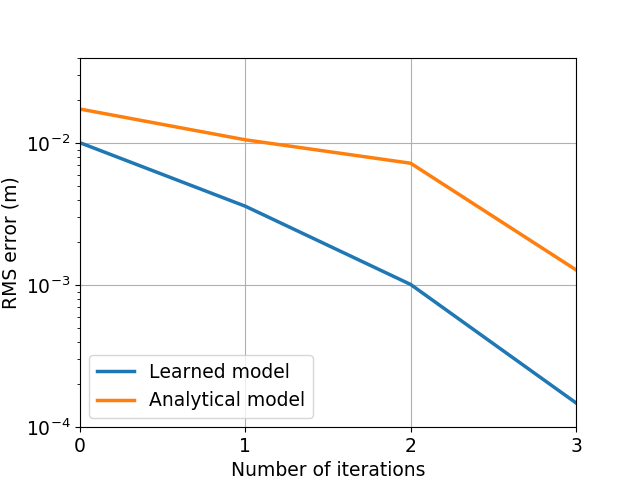}
% 		\begin{picture}(0,0)
% 	        \centering
% 	        \put(-185,150){Avg. Obj. Displacement = $0.043 \pm 0.033$ m}
% 		\end{picture}
	\end{subfigure}
	\begin{subfigure}[b]{0.24\textwidth} %\label{fig:multi_object_1}
		\centering
	\includegraphics[scale=0.5, angle=90]{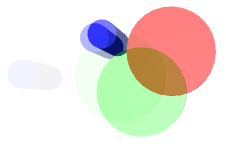}
	\begin{picture}(0,0)
		\centering
	        \put(-85,130){Analytical coarse model}
	        \put(-82,120){single-step prediction}
	\end{picture}
	\end{subfigure}
	\begin{subfigure}[b]{0.24\textwidth} %\label{fig:multi_object_2}
	    \centering
	\includegraphics[scale=0.5, angle=90]{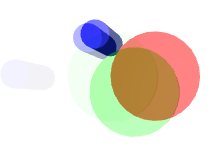}
	\begin{picture}(0,0)
	    \centering
	    \put(-80,130){Learned coarse model}
	    \put(-79, 120){single-step prediction}
	    \put(-140,20){\tikz\draw[red,fill=red] (0,0) circle (.5ex); Coarse prediction}
	    \put(-140,10){\tikz\draw[green,fill=green] (0,0) circle (.5ex); Fine prediction}
	    \put(-140,0){\tikz\draw[blue,fill=blue] (0,0) circle (.5ex); Pusher}
	\end{picture}
	\end{subfigure}
	\caption{Root mean square error (in log scale) of Parareal along the full trajectory for single object pushing using both a learned and an analytical coarse model (left).  {These results are for a control sequence with 4 actions where the average object displacement is $0.043 \pm 0.033~m$. The error at iteration four is 0.} The learned coarse model gives a better Parareal convergence rate. Sample motions for the learned coarse model (center) and the analytical coarse model (right). The learned coarse model's prediction is closer to the fine model prediction shown in green.}
	\label{fig:parareal_convergence_single}
\end{figure*}

In our experiments, we investigate three key issues.
First, we investigate how fast Parareal converges to the fine solution for robotic pushing tasks with different coarse models.
Second, we investigate the physics prediction accuracy of Parareal with respect to real-world pushing data.
Finally, we demonstrate that the Parareal physics model can be used to complete real-robot manipulation tasks.

In Subsection~\ref{sec:preliminaries} we provide preliminary information used throughout the experiments.
Subsection~\ref{sec:parareal_convergence} investigates convergence of Parareal for two different coarse models -- the analytical coarse model for single object pushing and a learned coarse model for both single and multiple object pushing.
In Subsection~\ref{sec:real_robot_experiments} we present results from real-robot experiments.
First, we compare the accuracy of Parareal predictions against real-world pushing physics.
Then, we show several real-robot plan executions using Parareal with a learned coarse physics model as predictive model.

\subsection{Preliminaries}
\label{sec:preliminaries}
{To generate physics-based robotic manipulation plans as fast as possible, we run Mujoco at the largest possible time-step (1ms) in all our experiments. Beyond this time-step the simulator becomes unstable, leading to unrealistically large object accelerations and breakdown of the simulator. We use the $4^{th}$ order Runge-Kutta integrator for Mujoco.}
All computations run on a standard Laptop PC with an Intel(R) Core (TM) i7-4712HQ CPU @2.3GHz with $N=4$ cores.
Our control sequences consist of four or eight actions, each applied for a control duration $\Delta_t = 1s$.

{The software version used to create training data and run experiments was Mujoco 2.00 with DeepMind DM Control bindings to Python 3.5 \cite{dm_control}. To develop, train and test the coarse model the Keras API was used, which is built in to TensorFlow 2.0. We used a learning rate of 5e-4 with 100 epochs and a batch size of 1024 to train the neural network model.
}

Our real robot setup is shown in Figure~\ref{fig:robot_manipulation}.
We have a Robotiq two-finger gripper holding the cylindrical pusher {of radius $1.45~cm$}. We place markers on the pusher and sliders to sense their full pose in the environment with an OptiTrack motion capture system.
{Sec.~\ref{sec:robotic_manipulation} states were defined to include orientation of objects but, to keep experiments simple, we use cylindrical objects such that only positions play a major role.}
{The slider radius used in all experiments is $5.12~cm$.}

\subsection{Parareal convergence}
\label{sec:parareal_convergence}
{Parareal produces the exact fine physics solution when the number of iterations is equal to the number of timeslices regardless of the coarse physics model~\cite{parareal_analysis_martin,parareal_Lions}. The convergence rate for scalar ordinary differential equations was theoretically shown to be superlinear on bounded intervals \cite{parareal_analysis_martin}.
However, for the differential algebraic equations in Eq.~\ref{eq:mujoco_equations} that describe the multi-contact dynamics problem, no such theoretical result exists and we study the convergence rate numerically.}

We investigate through experiments how fast Parareal converges using two coarse models - the analytic model for single object pushing and the learned model for both single object and multi-object pushing.
At each iteration, we compute a root mean square (RMS) error between Parareal’s predictions and the fine model's predictions of the corresponding sequence of states.
{We compute the RMS error over only positions since we used cylindrical objects in all experiments.}

\subsubsection{Single object pushing}
\label{sec:single_object_pushing_experiment}
% \begin{center}
% \begin{table}[b]
% \caption{Physics simulation time for a 4s long trajectory.}
% \centering
% \begin{tabular}{c c c}
% \hline\hline
%  & Simulation time (ms)\\
% \hline \hline
% Mujoco  & 124.4  \\
% Learned & 0.872  \\
% Analytical & 0.873   \\
% \end{tabular}
% \label{table:physics_simulation_time}
% \end{table}
% \vspace{-7mm}
% \end{center}
% %
% \vspace{-5mm}
We randomly sample an initial state for the pusher and slider.
We also randomly sample a control sequence where the pusher contacts the slider at least once during execution.
Thereafter, we execute the control sequence starting from the initial state using Parareal.
For the sample state and control sequence, we perform two runs, one using the learned model and the other using the analytical model as coarse propagator in Parareal.

We collect 100 state and control sequence samples. The analytical model makes a single step prediction 227.1 times faster than the fine model on average, while the learned model is 228.4 times faster on average. For example, to predict a 4s long trajectory, the fine model requires 1.22s while one iteration of Parareal requires only 0.31s (for both models) on average. We see that both coarse models are so fast that our actual speedup in using Parareal is almost completely governed by the number of iterations.

Furthermore, for these samples, we also compute the RMS error between Parareal and the fine model run in serial.
The results are shown in Fig.~\ref{fig:parareal_convergence_single} (left) for a control sequence with 4 actions where the average object displacement is $0.043 \pm 0.033~m$.

We see that the learned model leads to faster convergence of Parareal than the analytical model for single object pushing.
{One reason for this could be that, in general, more accurate coarse models lead to better convergence.}
The single-step prediction of the learned model, shown in read in Fig.~\ref{fig:parareal_convergence_single}~(right), is much closer to the fine prediction shown in green than the analytical model shown in ~Fig.~\ref{fig:parareal_convergence_single}~(center).

\begin{figure*}[htb!]
	\centering
	\begin{subfigure}[b]{0.49\textwidth}%\label{fig:multi_object_error}
	    \centering
		\includegraphics[scale=0.57]{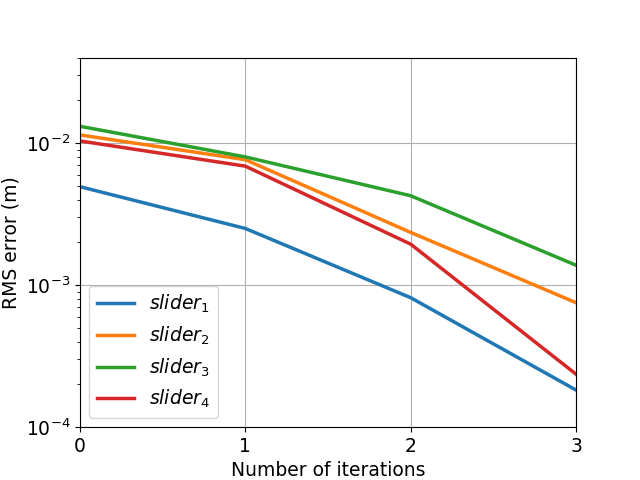}
% 		\begin{picture}(0,0)
% 	        \centering
% 	        \put(-185,157){Avg. Obj. Displacement = $0.015 \pm 0.029$ m}
% 		\end{picture}
	\end{subfigure}
	\begin{subfigure}[b]{0.24\textwidth} %\label{fig:multi_object_1}
		\centering
		\includegraphics[scale=0.37, angle=90]{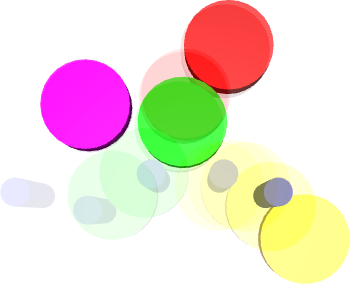}
		\begin{picture}(0,0)
		\centering
	        \put(-80,150){4-slider}
	        \put(-100, 140){Parareal prediction}
		\end{picture}
	\end{subfigure}
	\begin{subfigure}[b]{0.24\textwidth} %\label{fig:multi_object_2}
	    \centering
		\includegraphics[scale=0.39, angle=90]{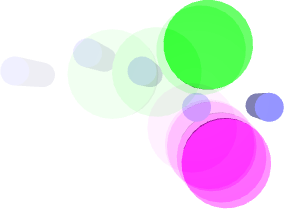}
		\begin{picture}(0,0)
	        \centering
	        \put(-65,150){2-slider}
	        \put(-85, 140){Parareal prediction}
		\end{picture}
	\end{subfigure}
	\caption{Root mean square error (in log scale) along the full trajectory per slider in a 4-slider pushing experiment (left) using \textit{only} the learned model. Two sample motions are illustrated (center and right) for multi-object physics prediction. {These results are for a control sequence with 4 actions where the average object displacement is $0.015 \pm 0.029~m$. The error at iteration four is 0 except for accumulation of round-off errors.}We find that the learned model enables Parareal convergence for the multi-object case}
	\label{fig:parareal_convergence_multi}
\end{figure*}
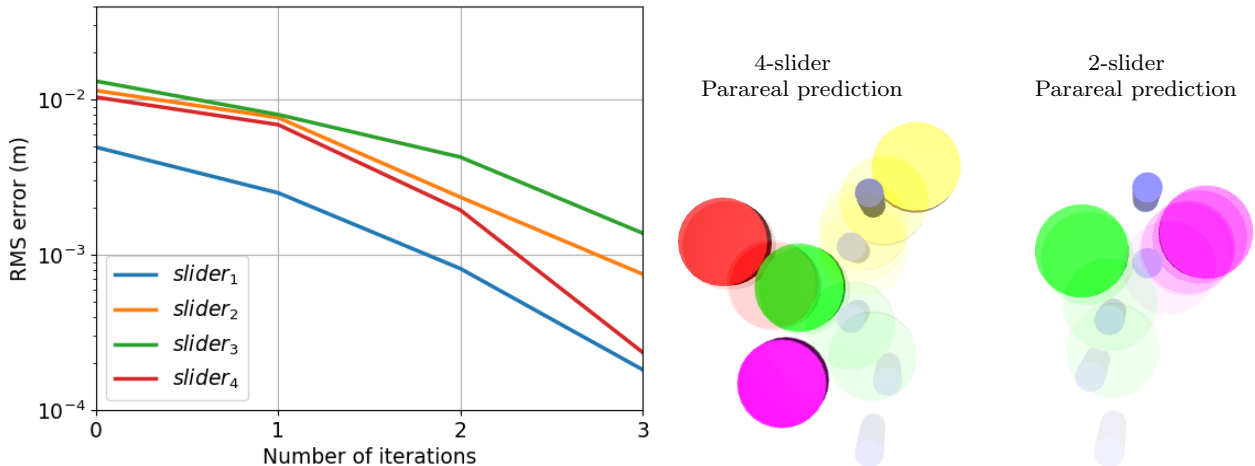

\subsubsection{Multi-object pushing}
\label{sec:multi_object_pushing_experiment}
We randomly sample a valid initial state for the pusher and multiple sliders.
Then, similar to the single object pushing case, we also sample a random control sequence that makes contact with at least one slider.
We then predict the corresponding sequence of states using Parareal.
However, for multi-object pushing we use only the learned model as the coarse physics model within Parareal.
The analytical model for single-object pushing would need significant modifications to work for the multi-object case.
Again, we collect 100 state and control sequence samples and run Parareal for each of them.
Our results are shown in Fig.~\ref{fig:parareal_convergence_multi}.

Fig.~\ref{fig:parareal_convergence_multi} (left) shows the RMS error per slider for each Parareal iteration.
While there are differences in the accuracy of the predictions for different slides, all errors decrease and Parareal converges at a reasonable pace.

{These results are for a control sequence with 4 actions and where average object displacement is $0.015 \pm 0.029~m$.} Some sample predictions are shown for a 4 slider environment in Fig.~\ref{fig:parareal_convergence_multi}~(center), and for a 2-slider environment in Fig.~\ref{fig:parareal_convergence_multi}~(right).
In both scenes, the pusher moves forward making contact with multiple sliders and Parareal is able to predict how the state evolves.

\begin{figure*}[htb!]
	\centering
	\begin{subfigure}[b]{0.49\textwidth}%\label{fig:multi_object_error}
	    \centering
		\includegraphics[scale=0.57]{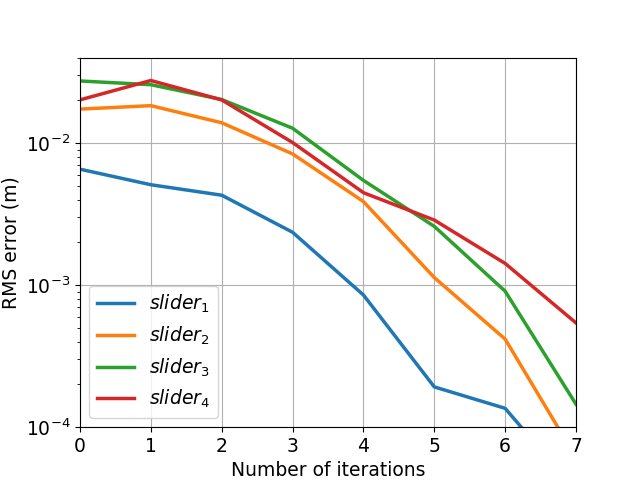}
% 		\begin{picture}(0,0)
% 	        \centering
% 	        \put(-185,157){Avg. Obj. Displacement = $0.034 \pm 0.082$ m}
% 		\end{picture}
	\end{subfigure}
	\begin{subfigure}[b]{0.49\textwidth}\label{fig:multi_object_error}
	    \centering
		\includegraphics[scale=0.57]{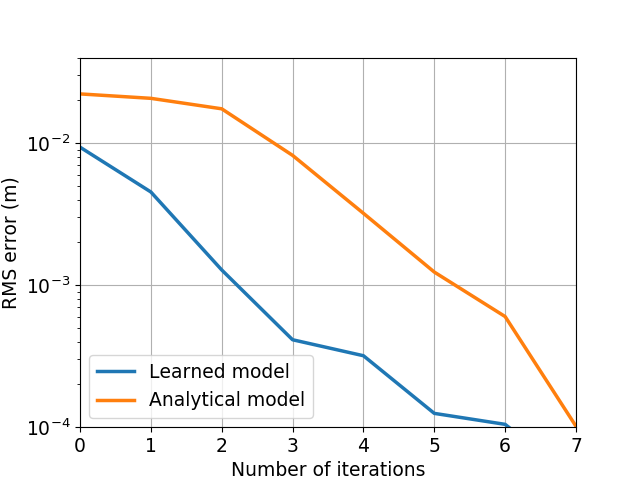}
% 		\begin{picture}(0,0)
% 	        \centering
% 	        \put(-185,157){Avg. Obj. Displacement = $0.046 \pm 0.040$ m}
% 		\end{picture}
	\end{subfigure}
	\caption{{Root mean square error (in log scale) along the full trajectory per object for single object pushing (right) and multiple object pushing(left) using \textit{only} the learned model. Here we consider a control sequence of 8 actions. The average object displacement for multi-object pushing is $0.034 \pm 0.082~m$ and for single object pushing it is $0.046 \pm 0.040~m$. The error at iteration eight is 0.
	 We find that the convergence of Parareal appears similar even with a longer control sequence.}
	 }
	\label{fig:parareal_convergence_8_steps}
\end{figure*}

{We also investigate Parareal convergence for a longer control sequence of 8 actions. We do this for single object and multi-object pushing where all other conditions are the same as for the 4-action control sequence. Results can be found in  Fig.~\ref{fig:parareal_convergence_8_steps}~(left) for multi-object pushing and  Fig.~\ref{fig:parareal_convergence_8_steps}~(right) for single object pushing. The average object displacement for multi-object pushing is $0.034 \pm 0.082~m$ and for single object pushing it is $0.046 \pm 0.040~m$. In general we find a similar convergence trend for both learned and analytical models for single and multi-object pushing.}

{Note that the shapes and sizes of the objects used are known and in fixed order. Therefore the learned model naturally does not generalize to new objects. However, it can still be used to make rather coarse predictions for similar objects.
}

\subsection{Real robot experiments}
\label{sec:real_robot_experiments}
\begin{figure}[t]
	\centering
	\includegraphics[scale=0.57]{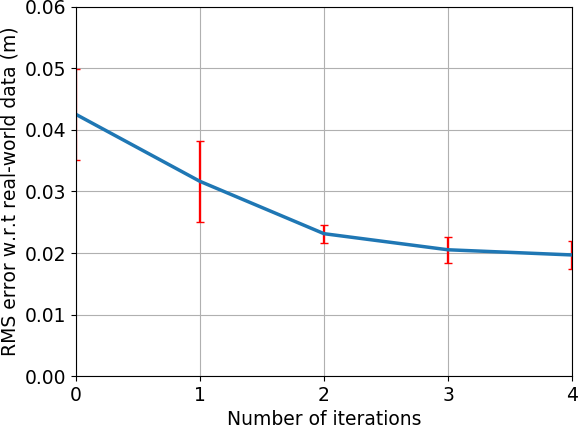}
	\caption{Root mean square error along the full trajectory for all 4 sliders \textit{measured with respect to the real-world pushing data}. The vertical bars indicate a 95\% confidence interval of the mean. The learned coarse physics model at iteration 0 has the largest error and the fine model provides the best prediction w.r.t the real-world pushing physics.}
	\label{fig:real_world_compare}
	\vspace{-5mm}
\end{figure}
In this section we investigate the physics prediction accuracy of Parareal with respect to real-world pushing physics. We do this for the multi-object case. In addition, we show real-world demonstrations for robotic manipulation where we use Parareal for physics prediction.

\subsubsection{Parareal prediction vs. real-world physics}
\label{sec:parareal_real_world}

Our coarse model neural network was trained using simulated data.
Here, we demonstrate that Parareal using the trained coarse model is also able to predict real-world states.
We randomly set an initial state in a real-world example by selecting positions for the pusher and sliders.
This state is recorded using our motion capture system.
Next, we sample a control sequence and let the real robot execute it.
Again, we record the corresponding sequence of states using motion capture.
Then, for the recorded initial state and control sequence pair, we use Parareal to produce the corresponding sequence of states and compare the result against the states measured for the real robot with optical tracking.

Figure~\ref{fig:real_world_compare} shows the RMS error between Parareal's prediction at different iteration numbers and the real-world pushing data.
Vertical red bars indicate 95\% confidence intervals.

Parareal's real-world error decreases with increasing iteration numbers and it is eventually twice as accurate as the coarse model.
These results indicate that Parareal's predictions with a learned coarse model are indeed close to the real-world physics predictions.
Figure~\ref{fig:real_world_planning_and_control} shows snapshots of the experiments.

\subsubsection{Planning and control}
\label{sec:planning_and_control}
\begin{figure*}[htb!]
  \centering
  \begin{subfigure}[b]{0.245\textwidth}
  \centering
     \includegraphics[scale=0.100]{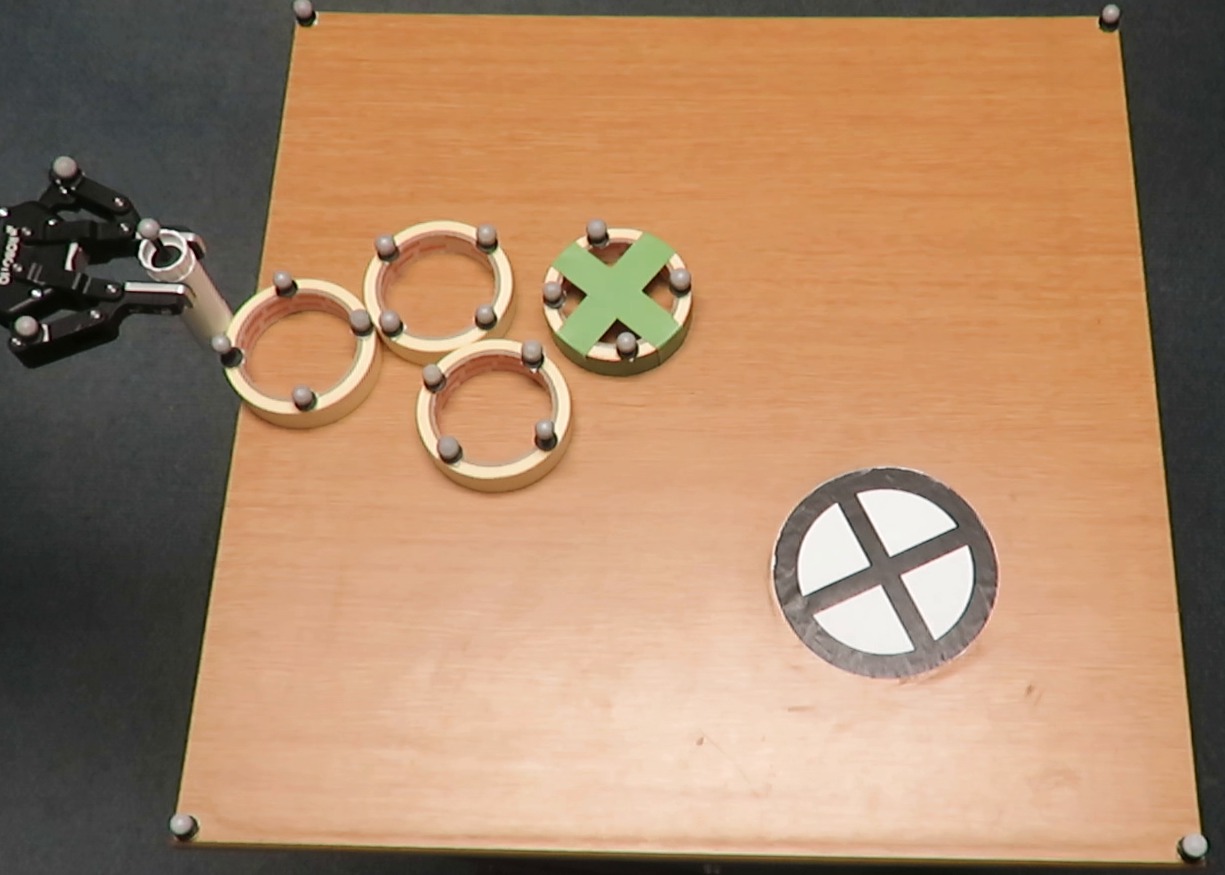}
    \end{subfigure}
  \begin{subfigure}[b]{0.245\textwidth}
  \centering
    \includegraphics[scale=0.100]{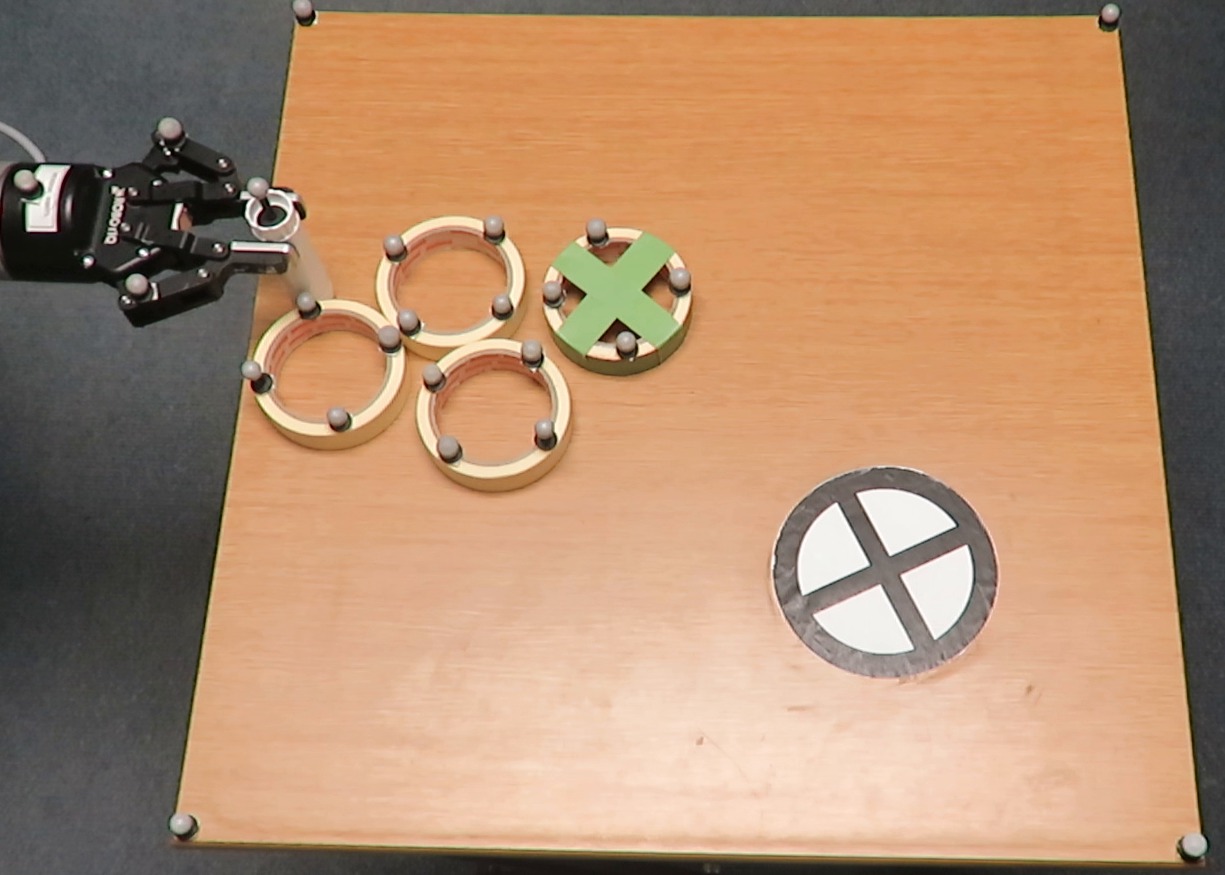}
  \end{subfigure}
     \begin{subfigure}[b]{0.245\textwidth}
    \includegraphics[scale=0.100]{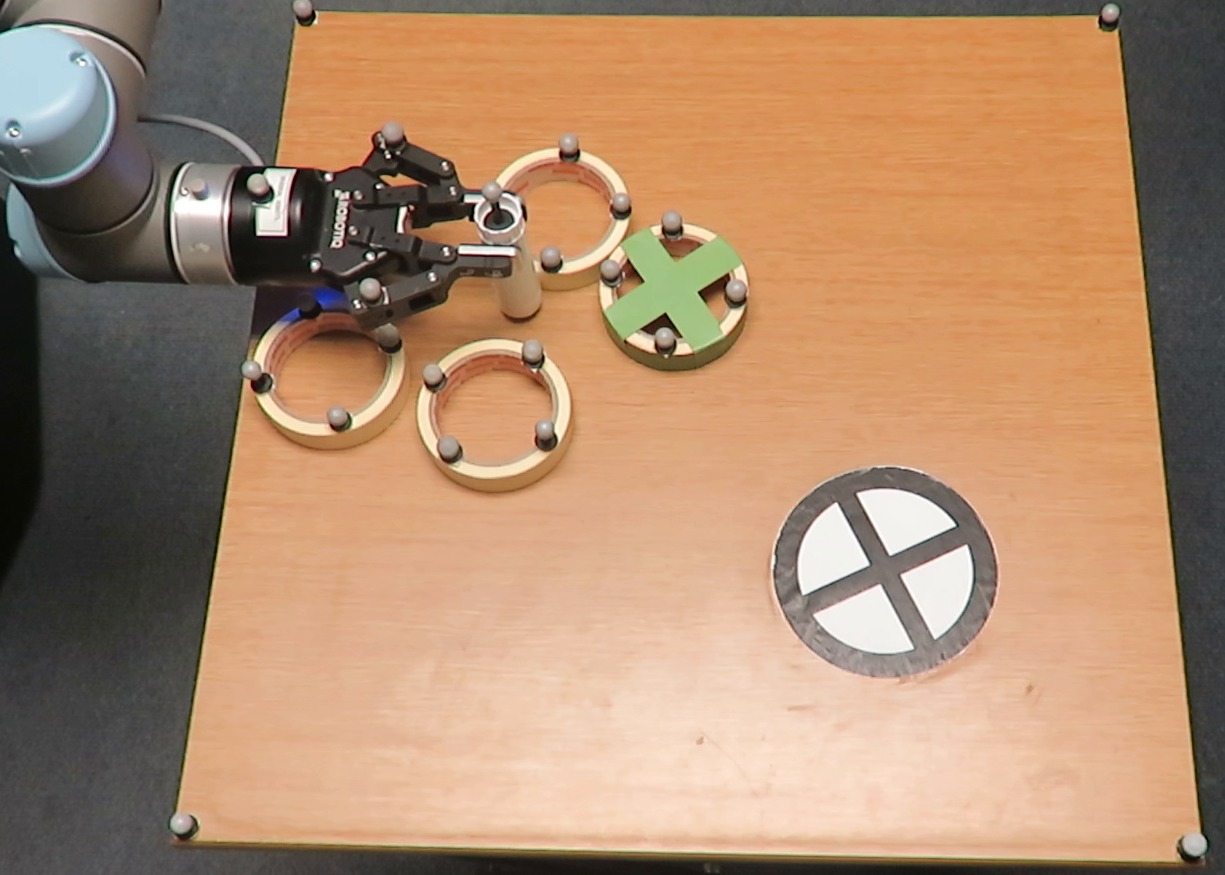}
  \end{subfigure}
  \begin{subfigure}[b]{0.245\textwidth}
    \includegraphics[scale=0.100]{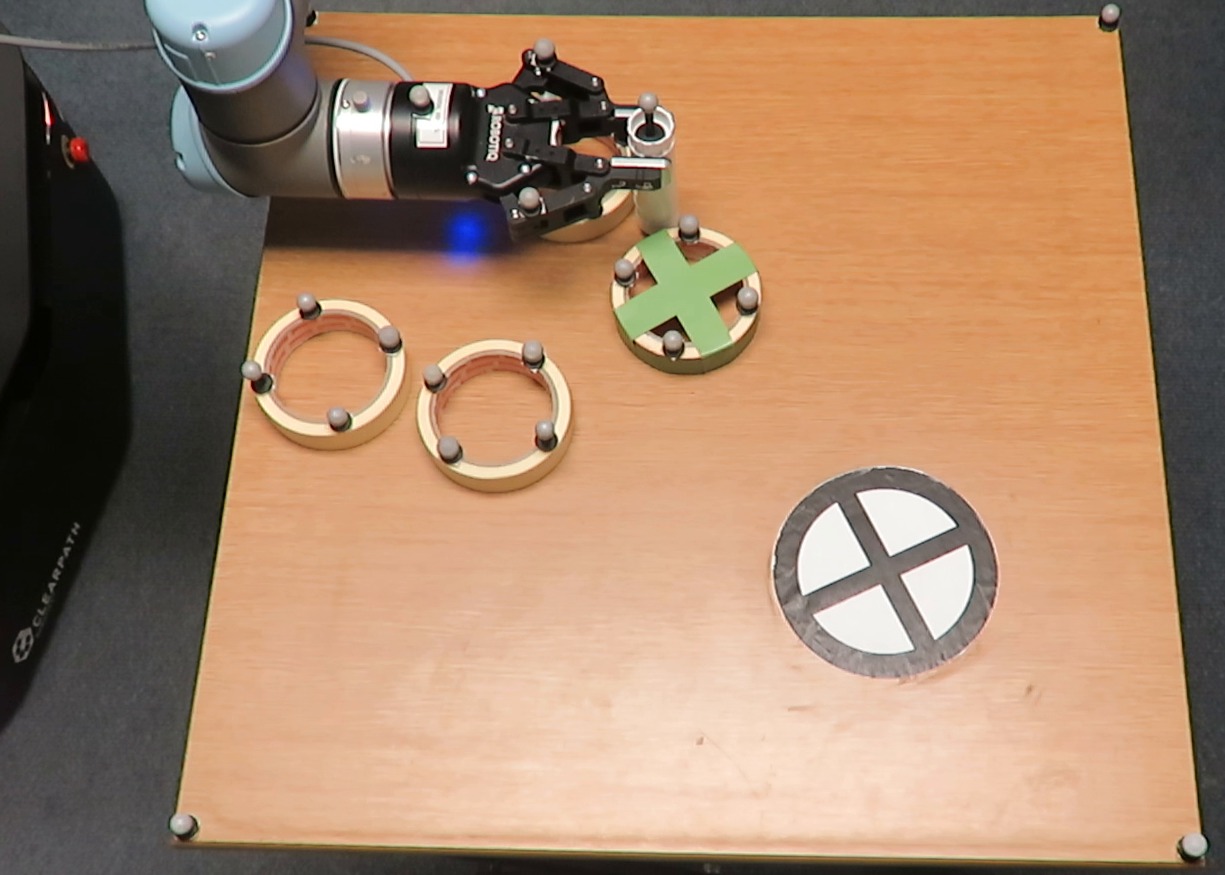}
  \end{subfigure}
  %%%%%%
  \begin{subfigure}[b]{0.245\textwidth}
  \centering
     \includegraphics[scale=0.100]{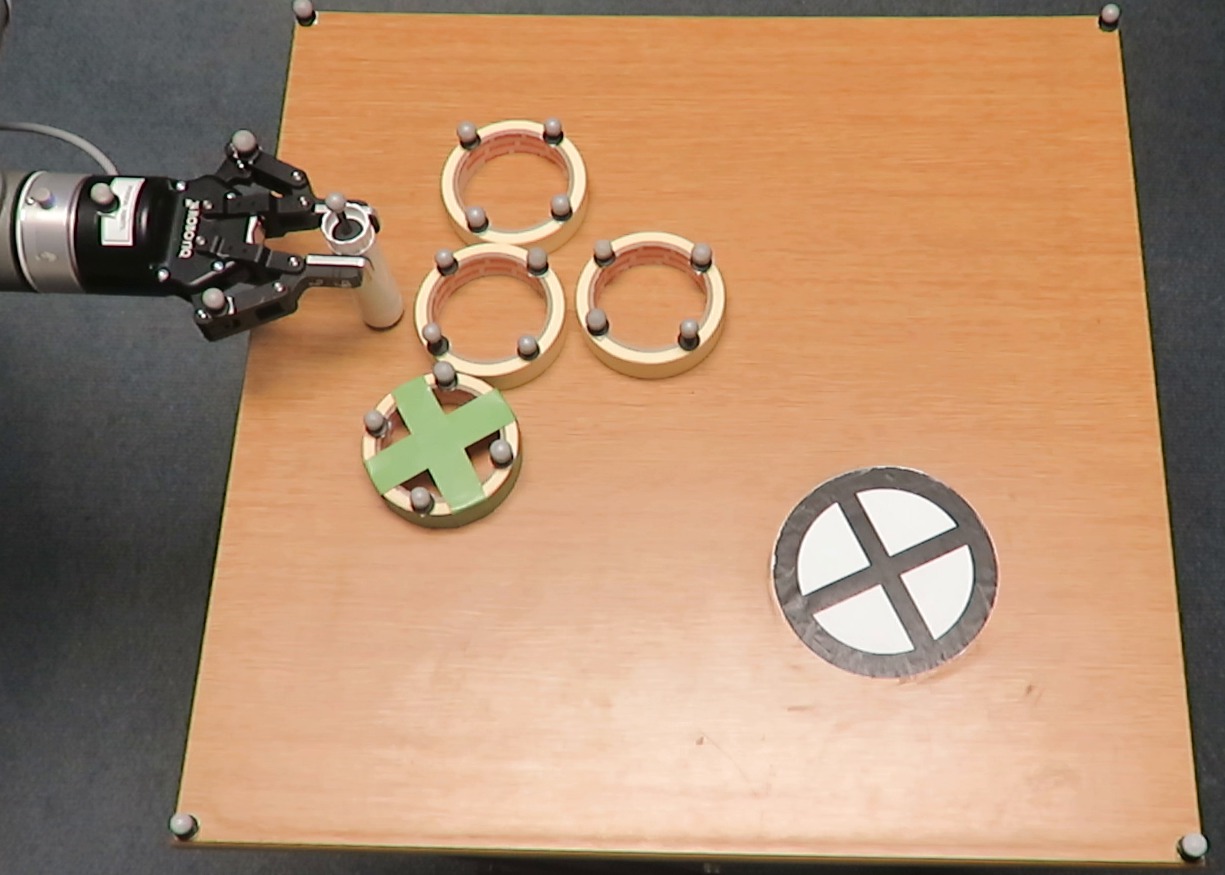}
    \end{subfigure}
  \begin{subfigure}[b]{0.245\textwidth}
  \centering
    \includegraphics[scale=0.100]{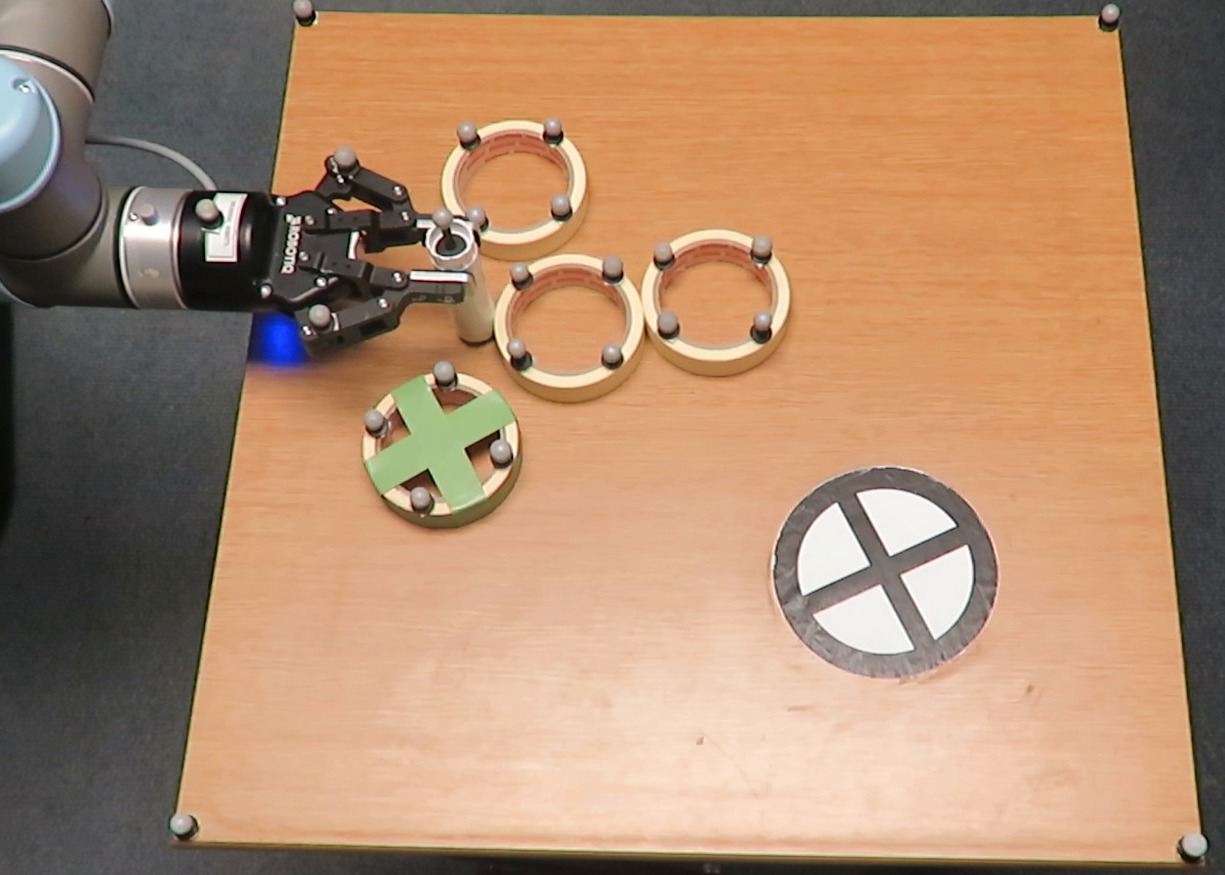}
  \end{subfigure}
     \begin{subfigure}[b]{0.245\textwidth}
    \includegraphics[scale=0.100]{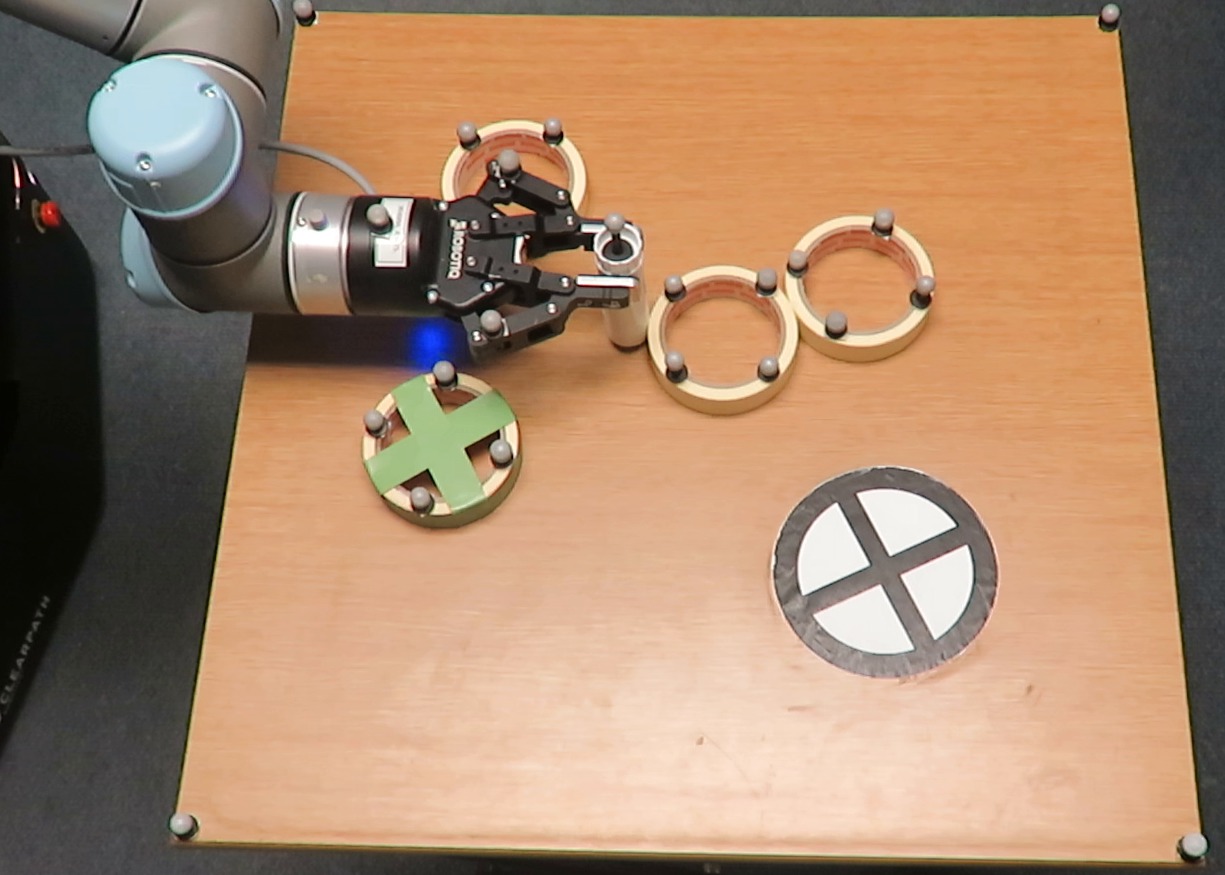}
  \end{subfigure}
  \begin{subfigure}[b]{0.245\textwidth}
    \includegraphics[scale=0.100]{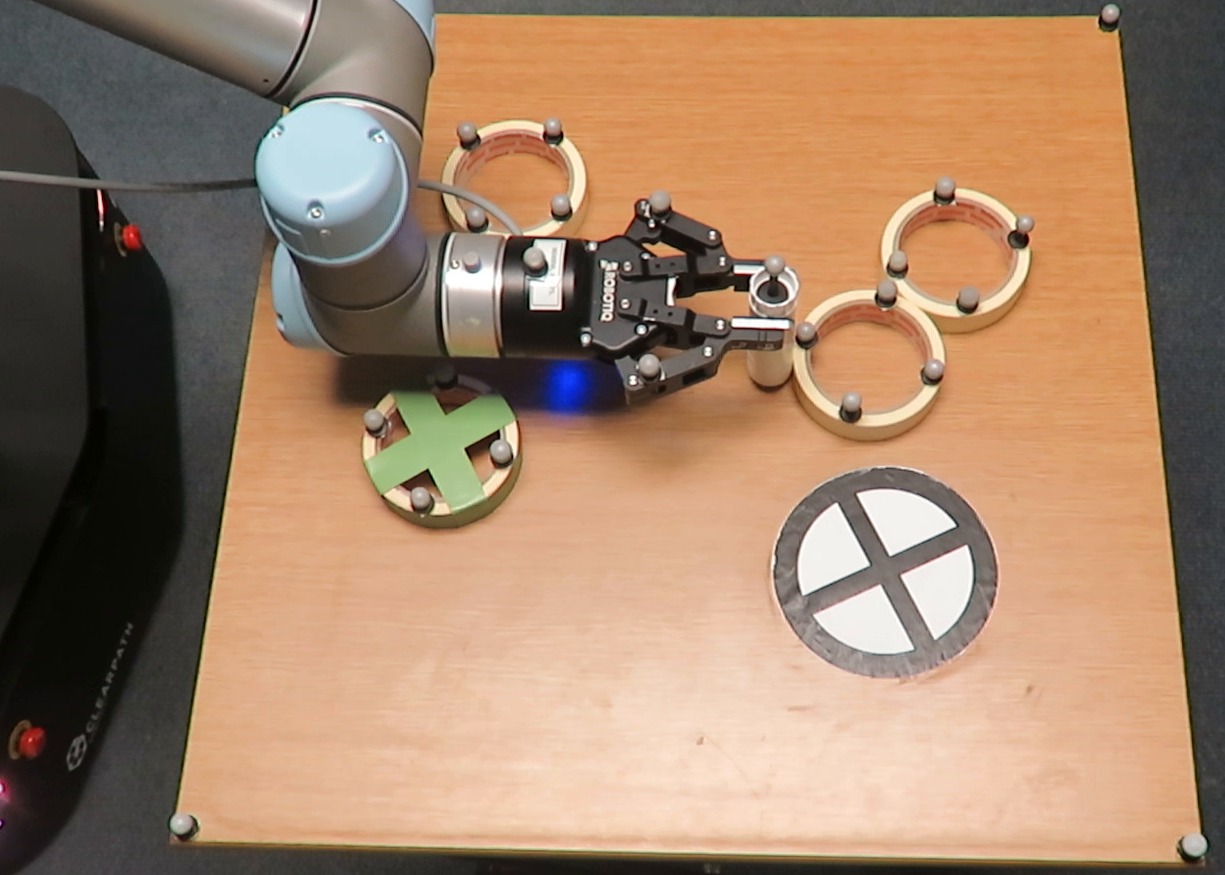}
  \end{subfigure}
%%%%%
  \begin{subfigure}[b]{0.245\textwidth}
  \centering
     \includegraphics[scale=0.100]{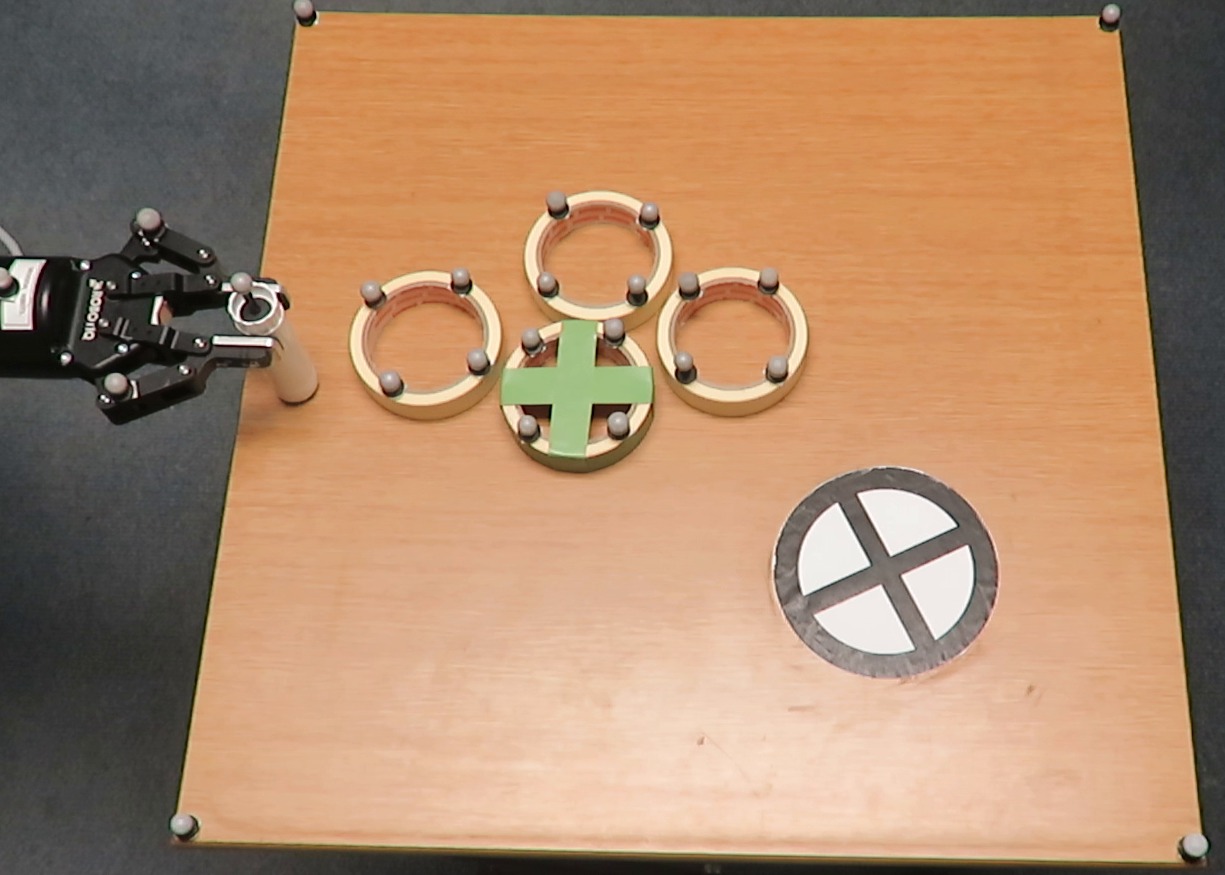}
    \end{subfigure}
  \begin{subfigure}[b]{0.245\textwidth}
  \centering
    \includegraphics[scale=0.100]{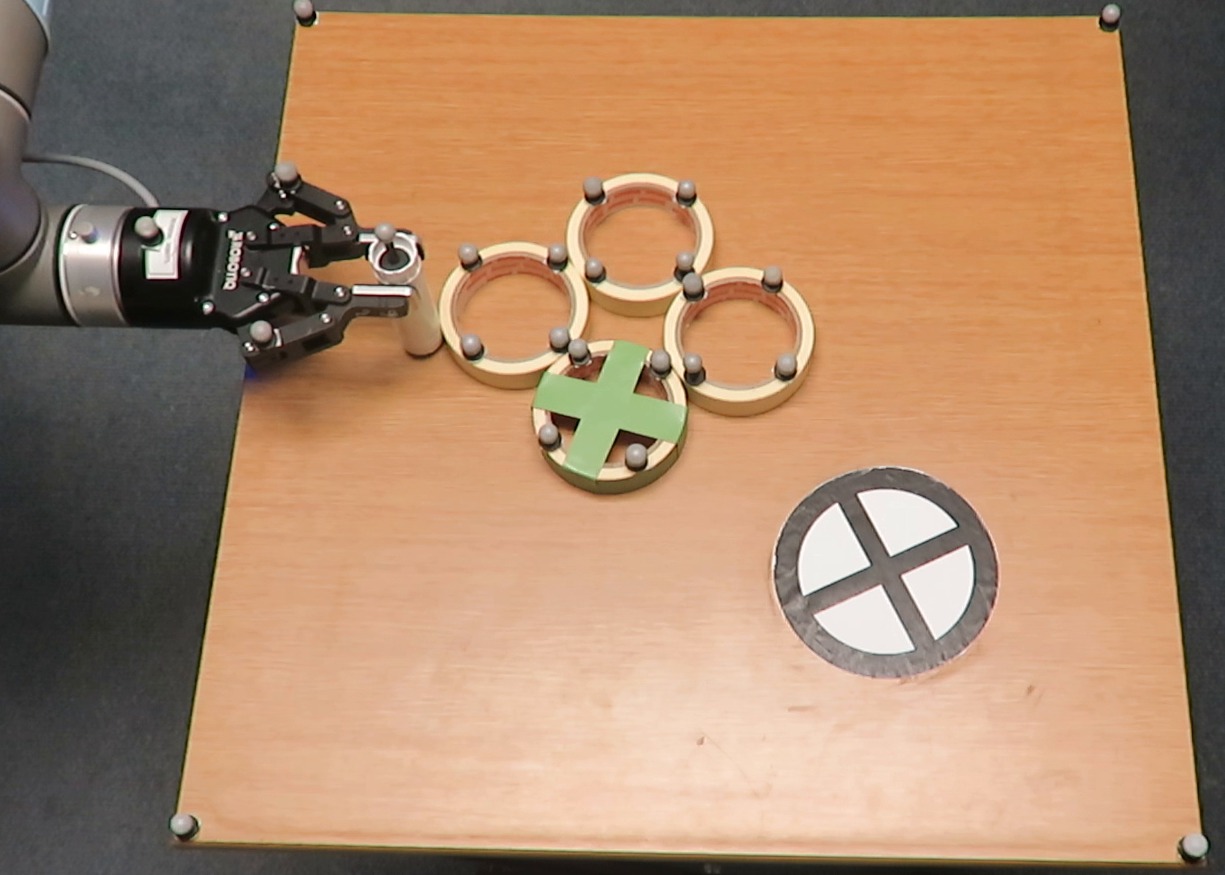}
  \end{subfigure}
     \begin{subfigure}[b]{0.245\textwidth}
    \includegraphics[scale=0.100]{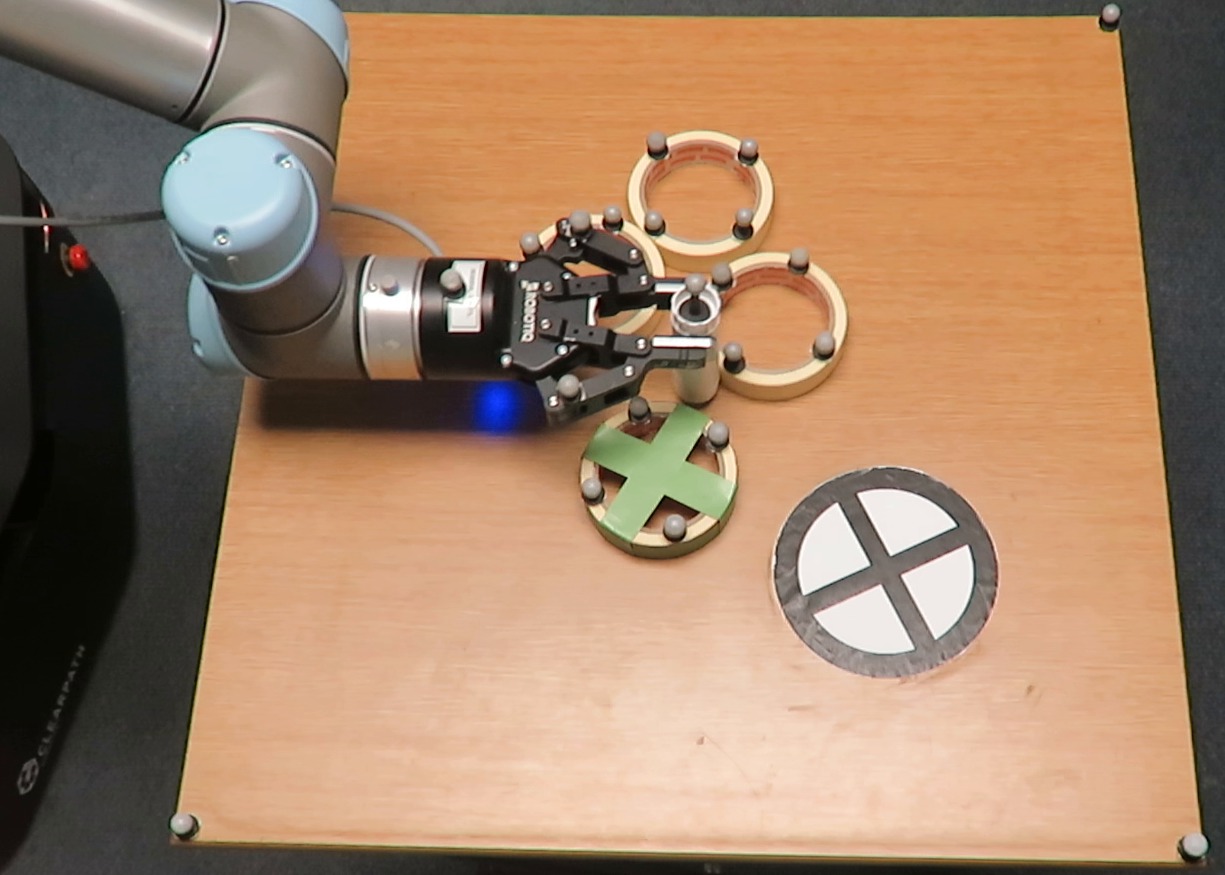}
  \end{subfigure}
  \begin{subfigure}[b]{0.245\textwidth}
\includegraphics[scale=0.100]{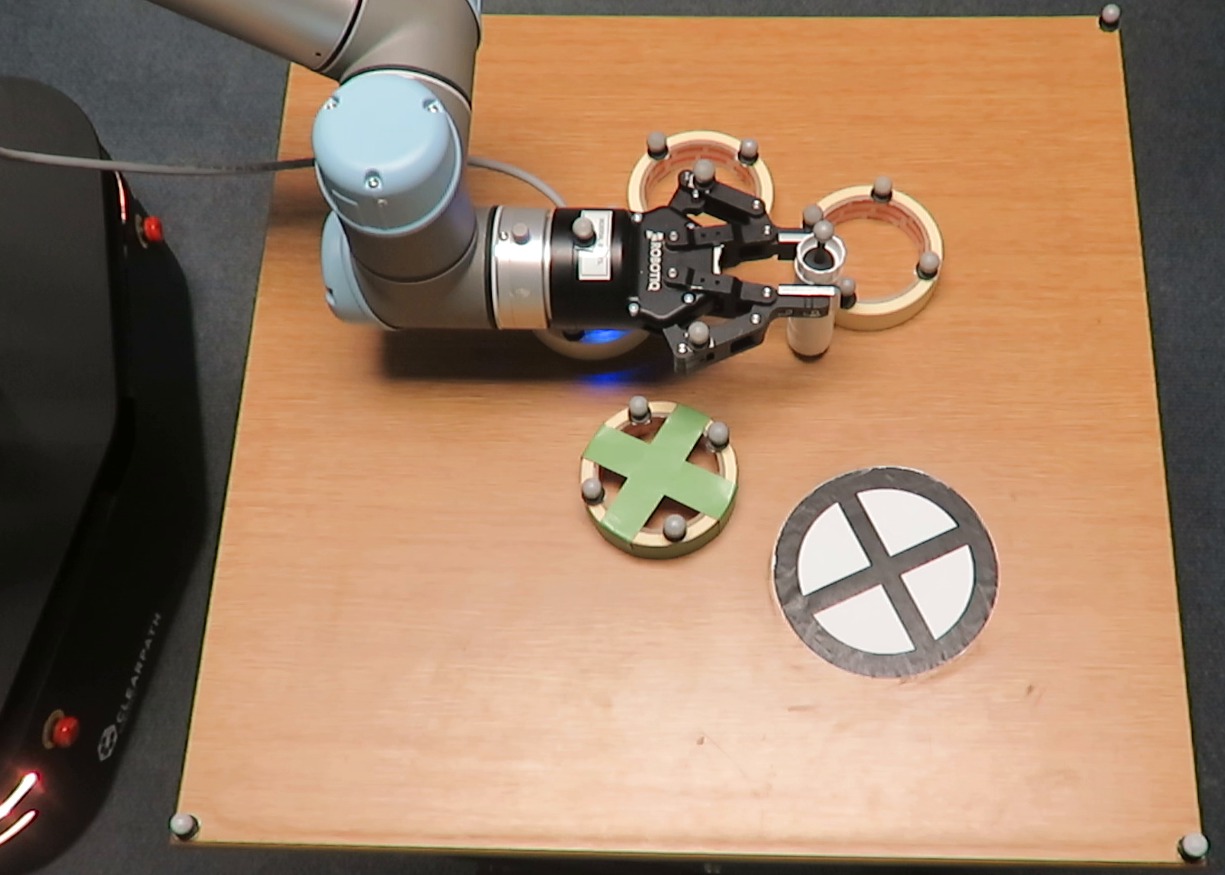}
  \end{subfigure}
\caption{The resulting sequence of states for applying a random control sequence starting from some random initial state in the real-world. Our goal is to assess the accuracy of the Parareal physics models with respect to real-world physics. We collect 50 such samples. These are some snapshots for 3 of such scenes - one per row with initial state on the left and final state on the right.}
\label{fig:real_world_snapshots}
\vspace{-3mm}
\end{figure*}
We use the Parareal predictive model for robotic manipulation to generate plans faster than using the fine model directly.
In this section, we complete 3 real robot executions with Parareal at 1 iteration.
We use the learned model as the coarse model in all cases.

\begin{figure*}[htb!]
  \centering
  \begin{subfigure}[b]{0.245\textwidth}
  \centering
     \includegraphics[scale=0.100]{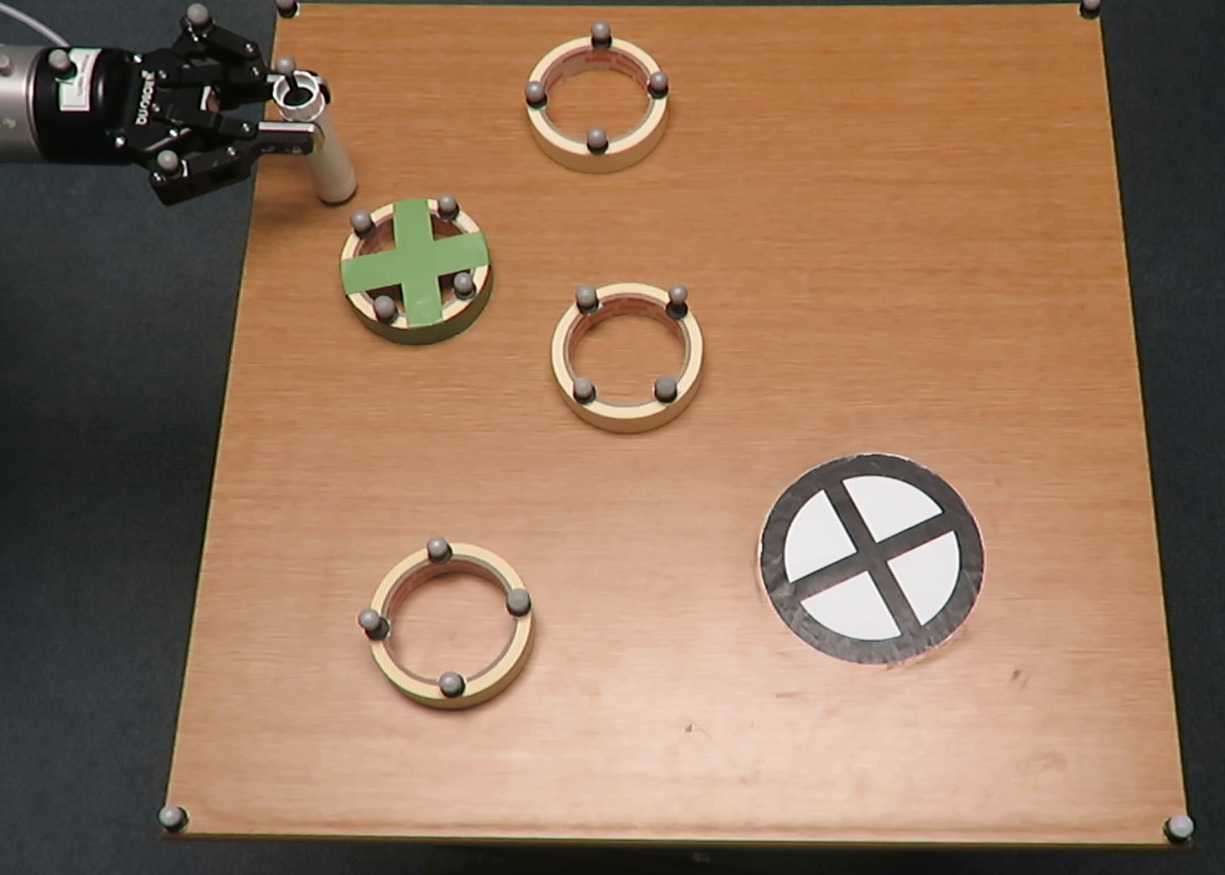}
    \end{subfigure}
  \begin{subfigure}[b]{0.245\textwidth}
  \centering
    \includegraphics[scale=0.100]{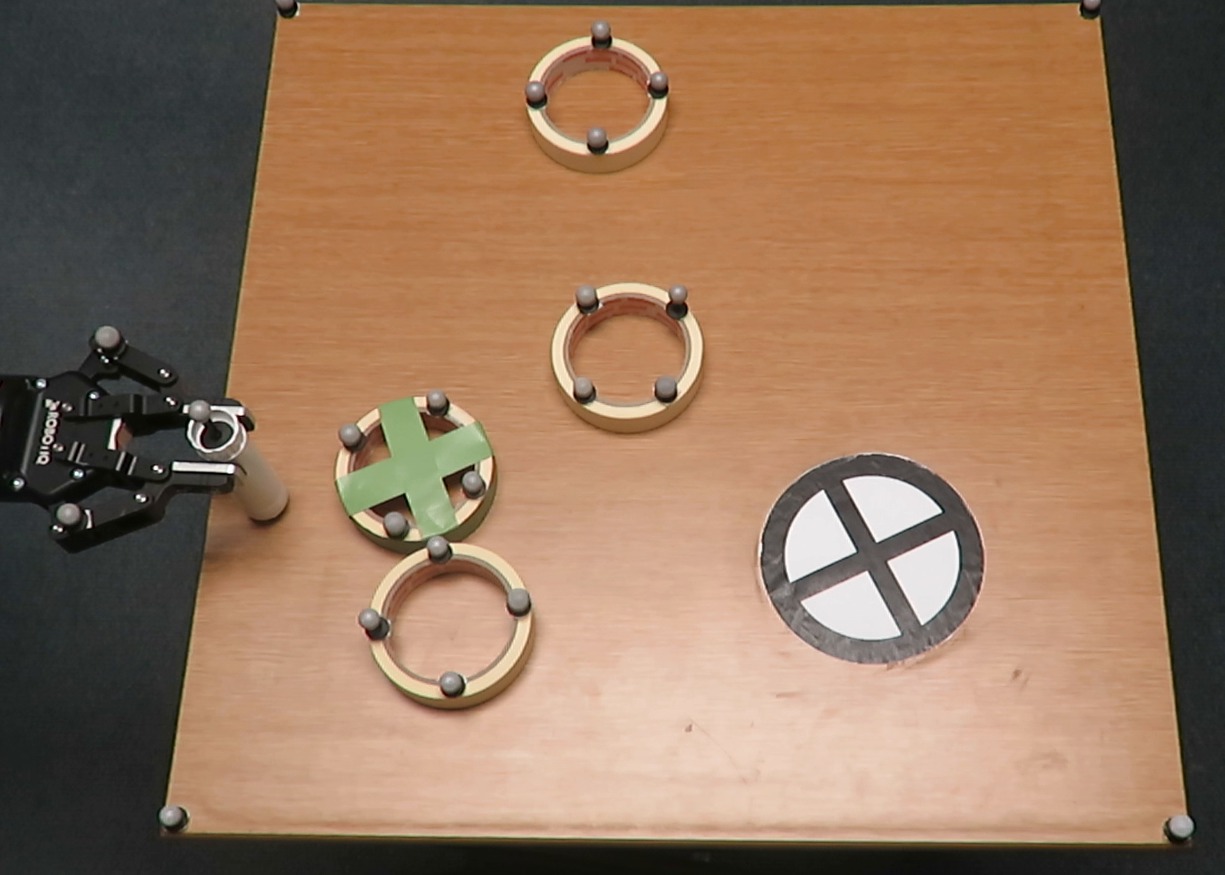}
  \end{subfigure}
     \begin{subfigure}[b]{0.245\textwidth}
    \includegraphics[scale=0.100]{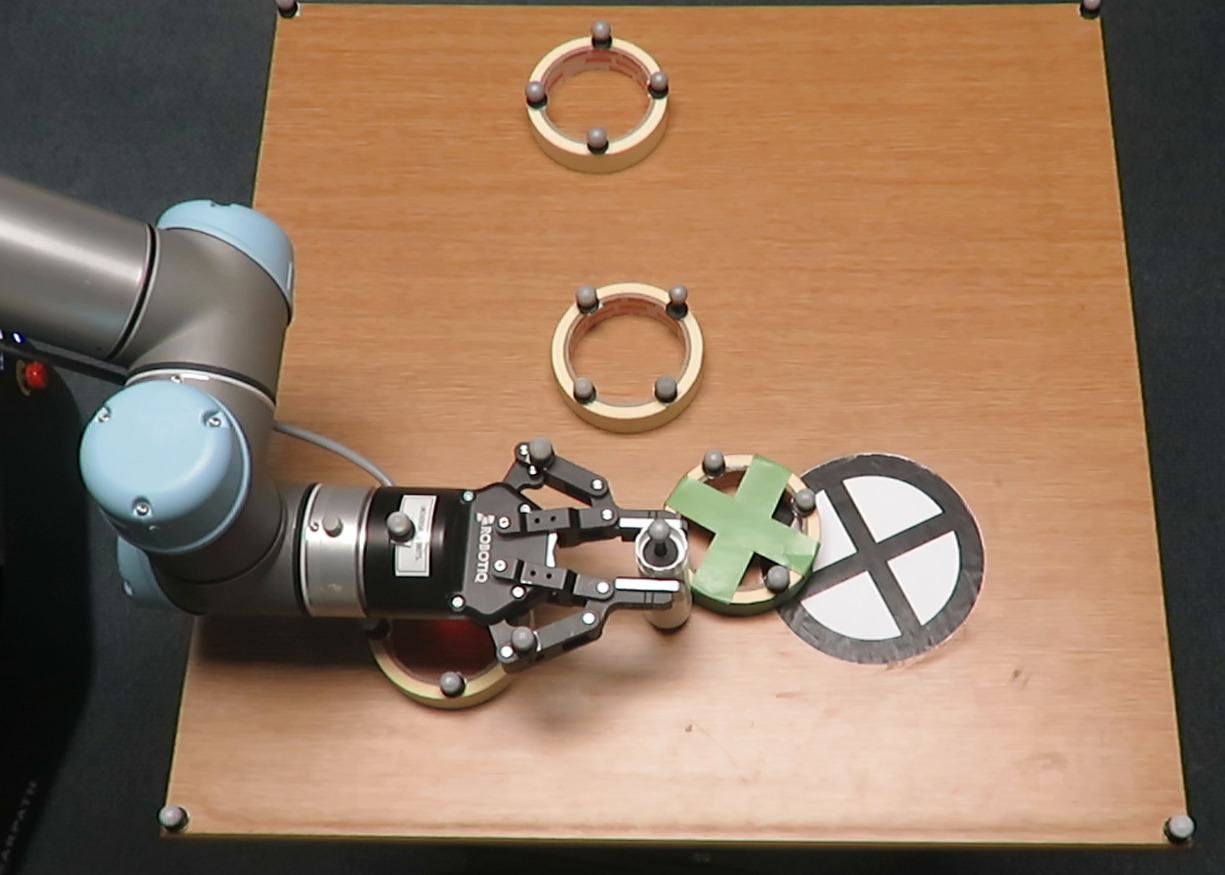}
  \end{subfigure}
  \begin{subfigure}[b]{0.245\textwidth}
    \includegraphics[scale=0.100]{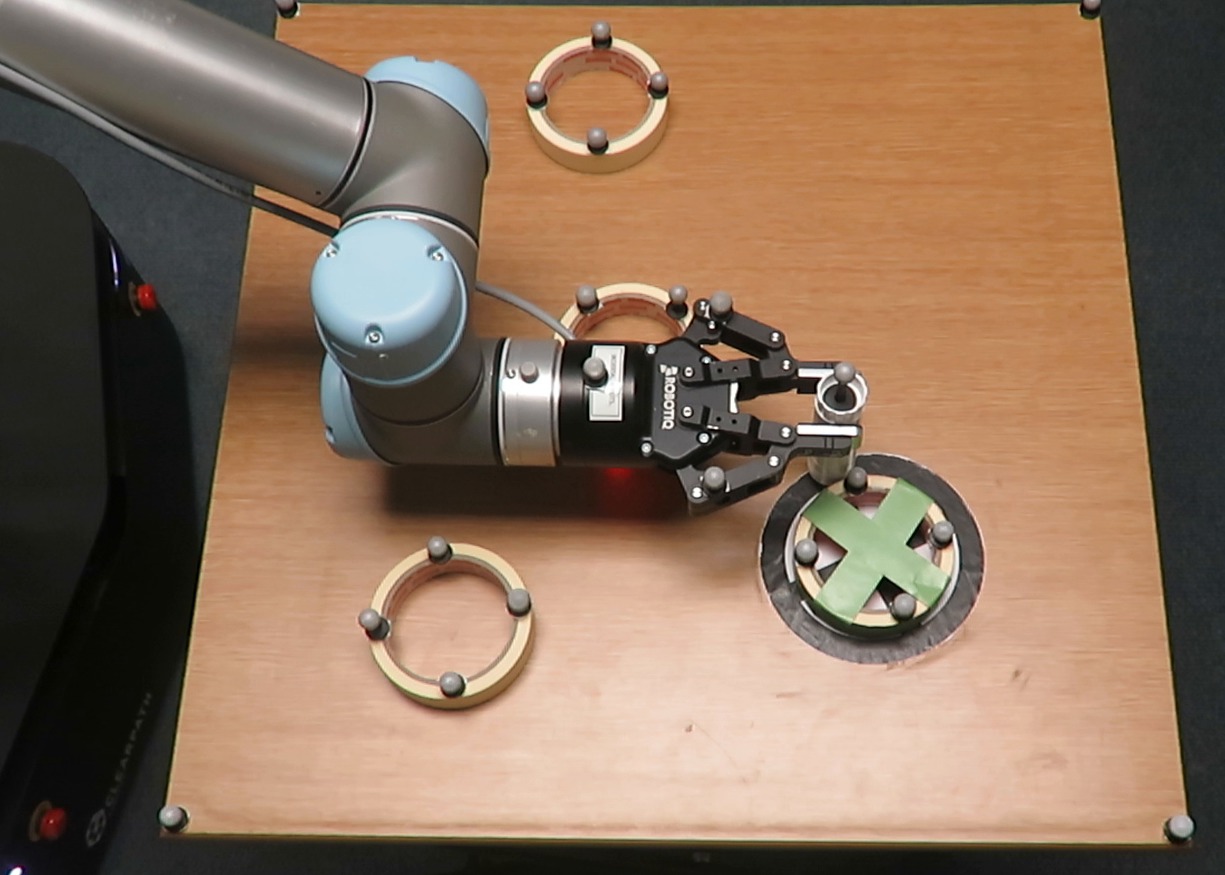}
  \end{subfigure}
%%%%
  \begin{subfigure}[b]{0.245\textwidth}
  \centering
     \includegraphics[scale=0.100]{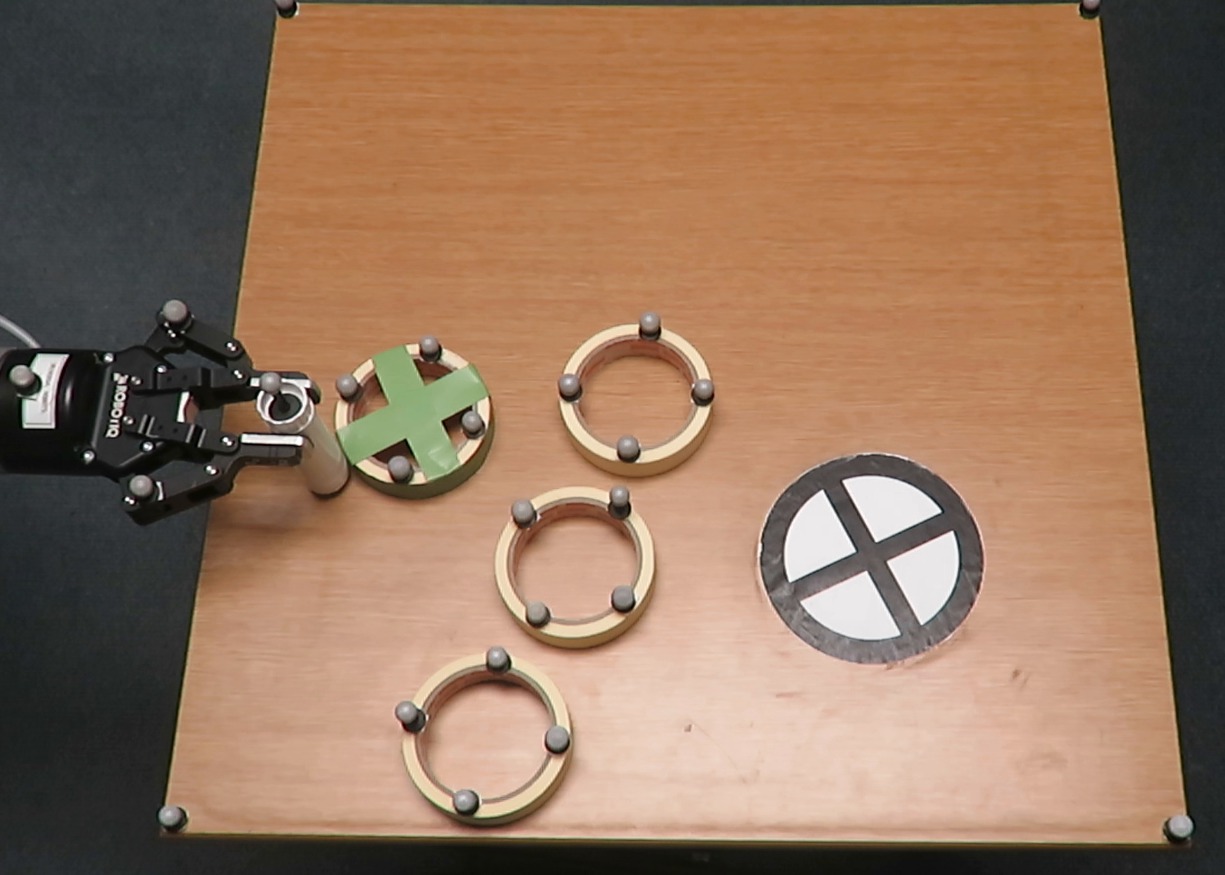}
    \end{subfigure}
  \begin{subfigure}[b]{0.245\textwidth}
  \centering
    \includegraphics[scale=0.100]{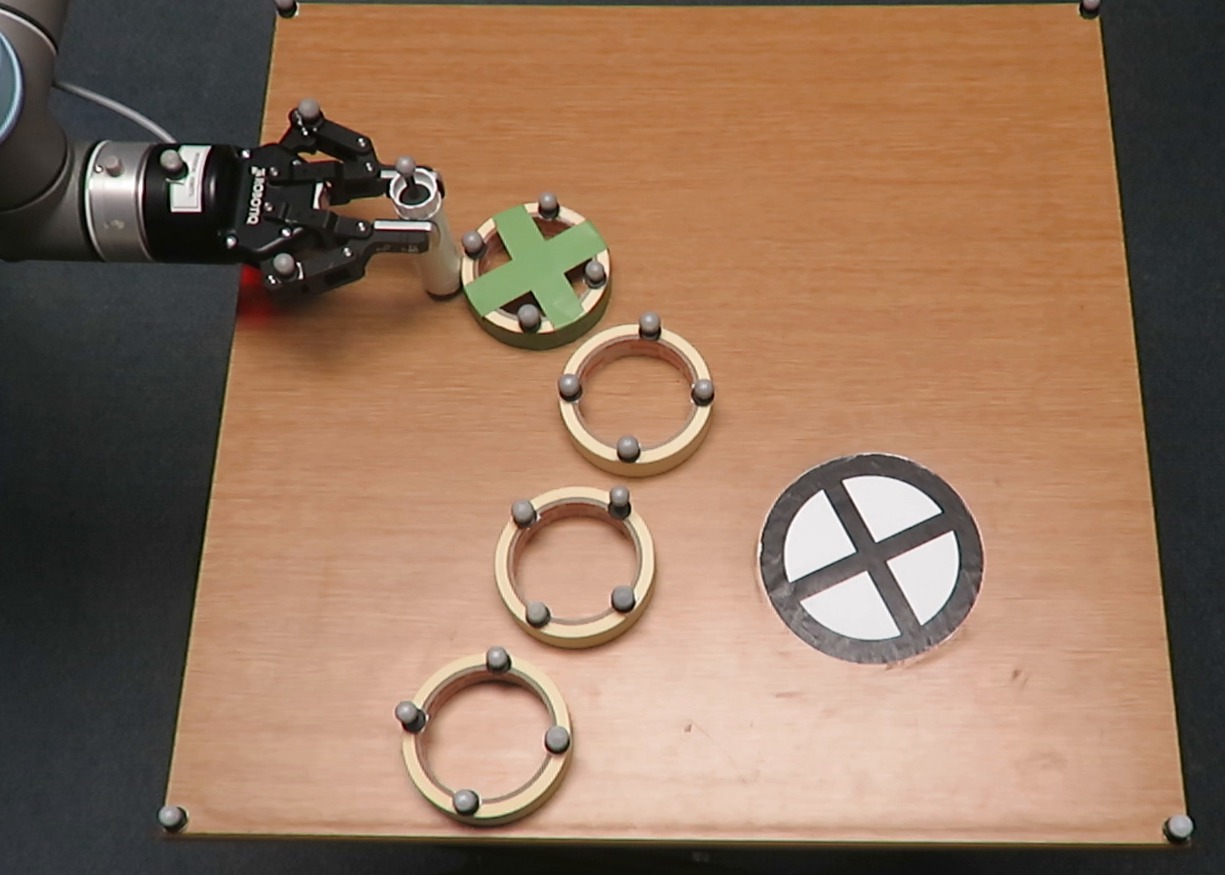}
  \end{subfigure}
     \begin{subfigure}[b]{0.245\textwidth}
    \includegraphics[scale=0.100]{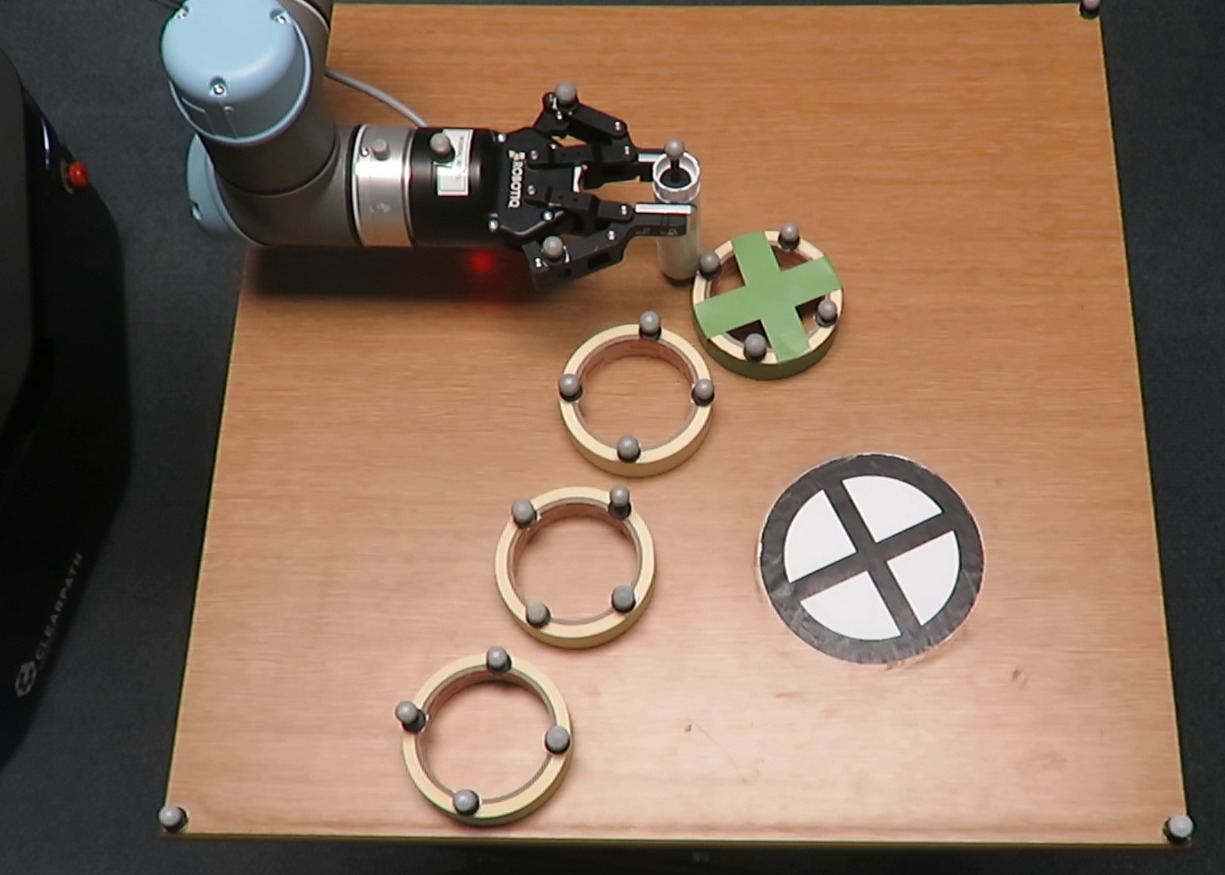}
  \end{subfigure}
  \begin{subfigure}[b]{0.245\textwidth}
    \includegraphics[scale=0.100]{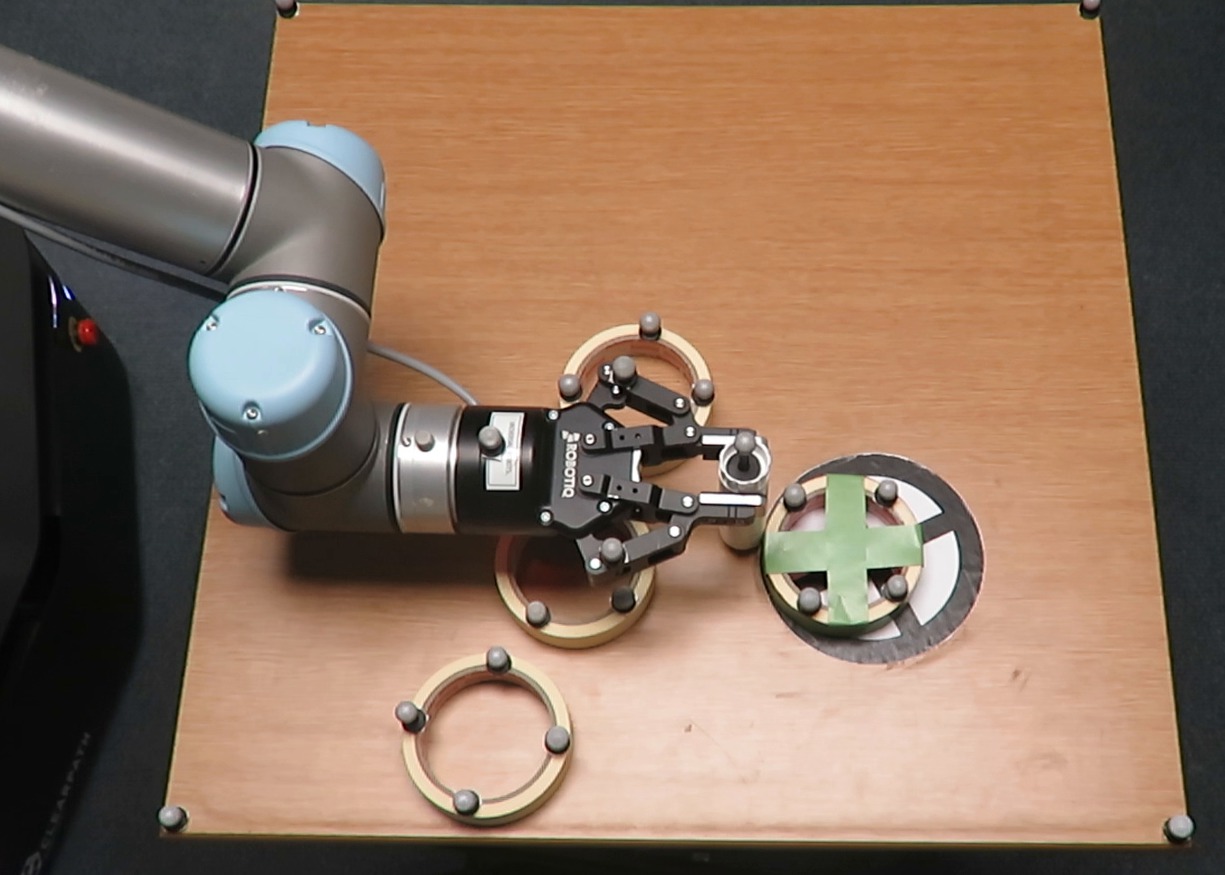}
  \end{subfigure}
\caption{Robotic manipulation planning and control for 2 different scenes.
The robot succeeds in all scenes using Parareal with a learned coarse model for physics predictions. The third planning and control scene is in Fig.~\ref{fig:robot_manipulation}.}
\label{fig:real_world_planning_and_control}
\end{figure*}

As can be seen in Figure~\ref{fig:real_world_planning_and_control}, the robot's task is to push the green slider into the target region marked with $X$.
The robot is allowed to make contact with other sliders. {An execution fails when a non-goal object is pushed into the goal region or over the edge of the table.}

The robot was successful for all 3 sample scenes. Some sample plans for two scenes are shown in Figure~\ref{fig:real_world_planning_and_control}. The third scene is shown in Figure~\ref{fig:robot_manipulation}.
We find that using Parareal with a learned coarse model for physics predictions, a robot can successfully complete complex real-world pushing manipulation tasks involving multiple objects.
At 1 Parareal iteration, we complete the tasks about 4 times faster than directly using the fine model.

In general, we trade-off physics prediction accuracy with respect to time. An important question then is how many iterations of Parareal to use for physics-based robotic manipulation i.e. how accurate should the physics predictions be? This depends on the manipulation task. For example, physics prediction accuracy should be higher when a robot is tasked with pushing an object on a narrow strip versus a large table where the chances of failure are lower.

Fig.~\ref{fig:real_world_compare} shows coarse physics errors (iteration 0) w.r.t. the real-world data of up to 5cm which is about the radius of a slider. Therefore, we conclude that the coarse model alone is not sufficient to complete the robotic manipulation task considered here --- an object can easily fall-off the table due to an inaccurately planned action.

Furthermore, there is uncertainty during robotic pushing in the real-world \cite{million_ways_to_push}. Agboh et. al. \cite{agboh_isrr19} showed that physics predictions with errors below real-world stochasticity (e.g. position standard deviation at the end of a real-world push) have similar planning success rates. Hence it is usually pointless to have physics predictions as accurate as the fine model.

\section{Summary}
We demonstrate the promise of using Parareal to parallelize the predictive model in a robot manipulation task involving multiple objects.
As coarse model, we propose a neural network, trained with a physics simulator.
We show that for single object pushing, Parareal converges faster with the learned model than with a coarse physics-based model we introduced in earlier work.
Furthermore, we show that Parareal with the learned model as coarse propagator can successfully complete tasks that involve pushing multiple objects.
We also show that although a simulator is used to provide training data, Parareal with a learned coarse model can accurately predict experiments that involve pushing with a real robot.

\bibliographystyle{spmpsci}
\bibliography{bib_file}

\end{document}